\pdfoutput=1
\documentclass[11pt]{article}

\usepackage[]{EMNLP2023} %review

\usepackage{times}
\usepackage{latexsym}

\usepackage[T1]{fontenc}
\usepackage[utf8]{inputenc}

\usepackage{microtype}

\usepackage{inconsolata}

\usepackage{graphicx} %Daniela
\usepackage{subcaption} %Daniela
\usepackage{paralist} %Daniela
\usepackage{diagbox} % Daniela

\newcommand{\DT}[1]{{\color{black} #1}} % Daniela
\newcommand{\DTtwo}[1]{{\color{black} #1}} % Daniela
\title{Language and Mental Health:\\ Measures of Emotion Dynamics from Text as Linguistic Biosocial Markers}

\author{Daniela Teodorescu$^{1,2}$\thanks{{$\;\;$Work done while at the University of Alberta.}}, Tiffany Cheng$^3$, Alona Fyshe$^{1,4}$,  Saif M. Mohammad{$^5$}\\
  $^1$Dept. Computing Science, Alberta Machine Intelligence Institute (Amii), University of Alberta\\
  $^2$MaiNLP, Center for Information and Language Processing, LMU Munich, Germany\\
  $^3$Carleton University \\
  $^4$Dept. Psychology, University of Alberta\\
  $^5$National Research Council Canada \\
  \texttt{\small \{dteodore,alona\}@ualberta.ca},
  \texttt{\small tiffany.cheng@carleton.ca}, 
  \texttt{\small saif.mohammad@nrc-cnrc.gc.ca}
  }

\begin{document}
\maketitle
\begin{abstract}

Research in psychopathology has shown that, at an aggregate level, the patterns of emotional change over time---emotion dynamics---are indicators of one's mental health. 
One's patterns of emotion change have traditionally been determined through self-reports of emotions; however, 
there are known
%these are known to have 
issues with accuracy, bias, and ease of data collection. %(e.g., time span and participant reach).
% It relies on self-reports of emotions, but recent work has explored emotion dynamics through textual utterances. This mitigates several limitations of self-report surveys (e.g., biases, limited reach). 
Recent approaches to determining emotion dynamics from one's everyday utterances addresses many of these concerns, but it is not yet known whether these measures of \textit{utterance emotion dynamics (UED)} correlate with mental health diagnoses. Here, for the first time, we study the relationship between tweet emotion dynamics and mental health disorders.
% at an aggregate level. 
We find that each of the UED metrics studied varied by the user's self-disclosed diagnosis. For example: average valence was significantly higher (i.e., more positive text) in the control group compared to users with ADHD, 
MDD, 
%depression,
and PTSD.
Valence variability was significantly lower in the control group compared to ADHD, depression, bipolar disorder, MDD, PTSD, and OCD but not PPD. Rise and recovery rates of valence also exhibited significant differences from the control. 
This work provides important early evidence for how linguistic cues pertaining to emotion dynamics can play a crucial role as biosocial markers for mental illnesses and aid in the understanding, diagnosis, and management of mental health disorders.

\end{abstract}

\section{Introduction}
% language is social, differences
Language is inherently \textit{social}---from the way in which we say things, the expressions we use and the things we choose to share, being impacted by our social environment and lived experiences.
% For centuries, language has been one of the primary means through which we communicate with each other and express ourselves.
%Language has been (for centuries) and continues to be one of the primary means through which we communicate with each other and express ourselves. 
%As our social environments have evolved over time, so has language to better support our communication needs and collaborative societies. 
As our social environments have evolved over time, language has evolved to better support our communication needs and collaborative societies.
%Therefore, language is also very \textit{variable}, as it has adapted to our cultures and communities around the world, and is influenced by one's experiences. 
Therefore, language is also \textit{variable}, as the way in which we use it has adapted to cultures and communities around the world, and it is influenced by an individual's experiences. 

%The prominent role of language in human evolution, and in our daily lives, also means that it is not surprising that our health impacts our language.
\DT{Given the prominent role of language in human evolution from hunters--gathers to collaborative societies, and the large extent to which we rely on language today, it is not surprising that our mental health impacts our language usage.}
% link between language and health biomarker
% Language plays an important role in our lives
% and communities, and also in
% %, from allowing us to communicate with each other and create collaborative environments, to 
% our health.
Quantitative features in language \DT{(e.g., aspects which can be measured)} have already been shown to indicate % predict 
and help \DT{clinicians} monitor the progression of mental health conditions (MHCs), acting as \textit{biomarkers}. 
% \DT{Linguistic biomarkers are quantitative features of language which have the potential to 
% %objectively
% predict the onset and progression of psychosis \cite{biosocial,gagliardi-tamburini-2022-automatic}.}
\DTtwo{A \textit{linguistic biomarker} is a language-based measure that is associated with %predicting 
a disease outcome or biology %of people 
in general \cite{biomarker,gagliardi-tamburini-2022-automatic,biosocial}.}
% for overall well-being and health. 
%Extracting quantitative linguistic features measures from text has been used for predicting the onset and progression of psychosis, and are often referred to biomarkers \citep{biosocial}. 
%For example, linguistic features such as:
\DT{Some well-known linguistic biomarkers include:}
the proportion of pronouns (indicator of depression,  \citet{koops}), syntax reduction (Anorexia Nervosa, \citet{Cuteri}), certain lexical and syntactic features (mild cognitive impairment and dementia, \citet{CALZA2021101113, Gloria}), and semantic connectedness (schizophrenia, \citet{CORCORAN2020158}. 
%  Linguistic Inquiry Word Count 
% categories differed in Twitter users with mental health disorders (ADHD, Generalized Anxiety Disorder, Bipolar Disorder, Borderline Personality Disorder, Depression, Eating Disorders, OCD, PTSD, schizophrenia, and seasonal affective disorder) \citep{coppersmith-etal-2015-adhd}, and more.
%Moreover,
\DT{Also,}
the emotions expressed in text have been shown to correlate with mental health diagnosis. For example, more negative sentiment 
\DT{in text by}
%in
individuals with depression \citep{DeChoudhury,seabrook,De_Choudhury_Gamon_Counts_Horvitz_2021}. Other work has shown that 
% in Reddit discourses that discuss mental health topics such as \textit{Suicide-Watch}, \textit{Anxiety}, and 
\DT{suicide watch, anxiety, and self-harm subreddits}
%self-harm subreddits 
had noticeably lower negative sentiment compared to other mental health subreddits such as Autism and Asperger's \citep{gkotsis-etal-2016-language}. 

% language is biosocial marker
%  true to varying degress, similar social groupings they have more stronger indicators of lang for health, some commonalities across social group, be careful of universality, better to use the term biosocial marker (lena),
While language can be a biomarker for mental health, the substantial social nature of language has implications. Notably, the tremendous variability in language use---especially across social groups---means that we should be skeptical about universal biomarkers; and instead realize that linguistic biomarkers alone are not capable of predicting MHCs. A vast amount of contextual and clinical information (often only available to 
\DT{an individual's}
%one's
physician) helps determine 
\DT{well-being},
%that,
\DT{and sometimes linguistic markers can aid the process.}
%sometimes with help from the linguistic markers.
Further, linguistic biomarkers are more likely to be a stronger indicator among groups with commonalities; for example, % such as 
%language, 
when applied to people from the same
region, culture, or medium of expression (e.g., social media platform). 
% As we have mentioned, language is very social and variable (e.g., lexical and syntactic features, emotion will differ based on language, upbringing, etc.), so we should be careful of the universality of such claims. 
%as there are many variations in language across backgrounds and cultures (e.g., syntactic and lexical features, emotion expressed).
%impacts such features like the emotions we express, and the structure of language.
%Although these trends have been shown across studies,
% While these trends appear in it is 
% While this may be true as similar social groups have more , 
%However, \citet{biosocial} propose the term \textit{biosocial} marker as linguistic markers are influenced by social aspects in addition to biological state.
For example, social factors such as parental socioeconomic status, neighbourhood, and institutionalization
\DT{(e.g., group foster care by government)}
influence speech markers such as lexical diversity; and social class influences syntactic complexity \citep{biosocial}. 
Therefore, it is more appropriate 
% term would be that 
to consider language as a \textit{biosocial} marker for health as it is influenced by both social  and biological factors \cite{biosocial}. 

% talk about the ethical implications since it is biosocial marker, can't look at some patterns doesn't mean it holds for every person
As language is increasingly being studied as a biosocial marker for mental health -- accelerated by the ease and availability of NLP tools and language data online -- there are important ethical implications.
%psychosis 
%it is important to
We must consider the sociolinguistic factors %of NLP markers for 
of such markers to ensure less biased and more accessible tools in clinics \citep{biosocial}. 
% For example, this includes being cautious of propagated biases from those found in training data and ensuring social diversity in participants \citep{biosocial}. 
If social factors are not considered, then this limits the utility of systems and their ability to predict well-being as they may be capturing confounding variables. 
% (e.g., semantic coherence is predictive of psychosis (a collection of symptoms that affect the mind, where there has been some loss of contact with reality), however individuals dropped out of school which is causing decreased semantic coherence rather than psychosis. %\footnote{Both dropping out of school and linguist features can contribute to psychosis, however there is decreased system utility \citep{biosocial}.}) 
% \citep{biosocial}. 

% our goal is to understand 
% Given this, 
In that context, our goal is to understand the extent to which patterns of emotion change 
act as
\textit{biosocial} markers for mental health? 
\textit{Emotion dynamics} studies the patterns with which emotions change across time 
% and involves the ``study of the trajectories, patterns, ... with which emotions ... fluctuate across time'' 
\cite{ed,KUPPENS201722}. 
Emotion dynamics have been shown to correlate with overall well-being, mental health, and psychopathology (the scientific study of mental illness or disorders) \citep{KUPPENS201722,Houben2015,depressedYouth,sperry}.
Further, studying emotion dynamics allows us to better understand emotions, and has been shown to have ties with academic success \citep{success, Phillips}, and social interactions \DT{(e.g., shyness)} in children \citep{Sosa}.

% Emotion Dynamics

% Mental health is key factor in overall well-being, and increasingly NLP techniques have been applied to increase knowledge and awareness in mental health research \citep{coppersmith-etal-2014-quantifying,hwang-hollingshead-2016-crazy,coppersmith-etal-2015-adhd,gkotsis-etal-2016-language,xu-etal-2020-inferring}. The way in which emotions change overtime -- \textit{emotion dynamics} -- correlate with overall well-being, mental health, and psychopathology (the scientific study of mental illness or disorders) \citep{KUPPENS201722,Houben2015,depressedYouth,sperry}.

%Specifically, emotion dynamics through various metrics quantifies features of the emotional episode (e.g., duration) and of the emotional trajectory (e.g., emotional variability, co-variation) \citep{KUPPENS201722}. 

Emotion dynamics have been measured in psychology through 
%ecological momentary assessments or 
self-report surveys over a period of time (e.g., 3 times a day for 5 days). Using these self-reports of emotion over time, various metrics can quantify how emotions change over time (e.g., the \textit{average intensity}, \textit{variability}, etc.).  %As there 
However, there are several drawbacks of using self-report surveys 
\DT{(which we discuss at the end of Section \ref{sec:emo_dyn})}.
%\DT{(limited amounts of data both in terms of duration and reach, and self-reports are prone to biases \cite{Kragel2022})},
Another window through which emotion dynamics can be inferred is through one's everyday utterances. 
\citet{movieED} proposed the \textit{Utterance Emotion Dynamics (UED)} framework which determines emotion dynamics from the emotion expressed in text. There are several metrics in the UED framework inspired by those in psychology (e.g., average emotion, emotion variability, rise rate, and recovery rate).
Ties between emotion dynamics metrics and mental health have been shown % extensively 
in psychology, however it is not known whether this relationship similarly exists between emotion dynamics in one's \textit{utterances/language} and mental health.

In this paper, we examine whether UED act as \textit{biosocial} markers for mental health. 
% and what differences or commonalities may exist between control and mental health diagnoses. %in the link  
X (formerly Twitter) provides an abundant amount of textual data from the public.\footnote{Since this work began before the name change, we use the terms \textit{Twitter} and \textit{tweets} in this paper.}
By considering tweets from users who have chosen to self-disclosed as having \DT{an}
MHC 
%mental health condition 
\citep{_Singh_Arora_Shrivastava_Singh_Shah_Kumaraguru_2022}, we can analyse the differences in UED metrics across a diagnosis and the control population.
%our work can be used to support other assessments by clinicians. 
%Due to the limited amount of mental health annotated data (due to privacy), we use a Twitter dataset of users who have self-disclosed as having a mental health condition \citep{_Singh_Arora_Shrivastava_Singh_Shah_Kumaraguru_2022}.
%We are interested in if emotion dynamics from utterances can act as a \textit{biosocial marker} \citep{biosocial} for mental health. 
%Our work provide important findings for clinicians by providing a broader context for overall well-being. 
%The primary research goal is to
% In this paper 
We describe how %we
%To examine how
utterance emotion dynamics (UED) metrics compare between different each of the seven diagnoses (ADHD, bipolar, depression, MDD, OCD, PPD, PTSD) and a control group; and for each MHC we explore the following research questions:\\[-20pt] %where each corresponds to an UED metric
% \begin{compactitem}
\begin{enumerate}
\item Does the average emotional state %across time
differ between the MHC and the control group?\\[-20pt]
\item Does emotional variability differ between the MHC and the control group?\\[-20pt]
\item Does the rate at which emotions reach peak emotional state (i.e., rise rate) differ between the MHC and the control group?\\[-20pt]
\item Does the rate at which emotions recover from peak emotional state back to steady state (i.e., recovery rate) differ between the MHC and the control group?\\[-20pt]
\end{enumerate}
% \end{compactitem}
\noindent We explore each of the above research questions for three \DT{dimensions of} emotion --- valence, arousal, and dominance --- further building on the findings in psychology which have traditionally focused only on valence. Our work provides baseline measures for UEDs across MHCs and insights into new linguistic biosocial markers for mental health. 
These findings are important for clinicians 
%by providing
\DT{because they provide}
a broader context for overall well-being and can help contribute to the early detection, diagnosis, and management of MHCs.
% However, it is important to note that we are not developing systems for `detecting' or `classifying' an individual's mental health status. There are many other factors to consider in addition to one's utterances, and there are ethical implications in trying to do so.
% Rather, we are interested in how patterns of emotion change may differ between groups at the aggregate level.
%Further,
% We note that 
% UED metrics should never be used as the sole determiners for mental health but rather in support of other assessments by clinicians.

% We, for %Our work, for the 
% the first time, demonstrate the significant relationships between UED and mental health diagnoses. 

% Hypothesis
% %move to just before methods?
% \begin{compactitem}
%     \item 
% \end{compactitem}

%Emotion arc??

\section{Background}
How people feel on average and the patterns of emotional change over time are well supported as a unique indicator of psychological well-being and psychopathology \citep{Houben2015}. 
Below we review related work on emotion dynamics and overall well-being in psychology and related fields.

\subsection{Emotion Dynamics}
\label{sec:emo_dyn}
Affective chronometry is a growing field of research that examines the temporal properties of emotions (i.e., emotion dynamics) \DT{and increasingly its} 
%in
relation to mental health and well-being \DT{have been studied} \citep{Kragel2022}. Within this field, emotion reactivity is the threshold at which a response is elicited and the responses' amplitude and duration - it can be further deconstructed into the rise time to the peak response (i.e., rise rate), the maintenance time of the peak response, and the recovery time to baseline (i.e., recovery rate) \citep{davidson1998}. 
Emotion variability is how variable an emotion is in terms of its intensity over time and in response to different stimuli. This variation can occur over multiple timescales (e.g., seconds, hours, days, weeks; \citealp{Kragel2022}). 
%Emotional inertia is the resistance to emotional changes over time, leading to carryover effects of emotional experience despite changes in the external and internal environment \citep{kuppens_Verduyn2017}. 
These emotion dynamic metrics are proposed to be predictive of affective trajectories over time and in distinguishing between affective disorders (\citealp{Kragel2022}). 
%As well as 
Also, emotion dynamics
contribute to maladaptive emotion regulation patterns and poor psychological health \citep{Houben2015}. % Notably, 
%Further, t
The meta-analysis by \citet{Houben2015} has indicated that the timescale of emotion dynamics does not moderate the relation between emotion dynamics and psychological well-being. 
\DT{Therefore, the relationship between psychological well-being and emotion dynamics occurs
whether it is examined over seconds, days, months, and so on.}
%Thus,
%the effect of psychological well-being on metrics of emotion dynamics is detectable whether it is examined over seconds, days, months, and so on. 
% Our study examines metrics of emotion dynamics across affective disorders including depression, major depressive disorder (MDD), and post-partum depression (PPD) as well as other psychopathologies (e.g., post-traumatic stress disorder, bipolar disorder, attention-deficit/hyperactivity disorder, obsessive-compulsive disorder).

\noindent {\it Average Emotional State \& Psychopathology:}
Average or baseline emotional states are related to well-being and mental illnesses. Due to the maladaptive (i.e., dysfunctional) nature of psychopathology, those with mental illnesses tend to have more negative emotional baselines. %compared to those without. 
For example, \citet{Heller2018} found that a higher average positive affect is associated with lower levels of depression but not anxiety, and a higher average negative affect was related to increased depression and anxiety. As well, those with post-traumatic stress disorder (PTSD) have reported lower average positive affect \citep{Pugach2023}. 
% Emotional baselines for attention-deficit/hyperactivity disorder (ADHD) have not been reported but ADHD has been proposed to lead to the tendency to avoid emotional variance and maintain affective neutrality \citep{adhdEmote} indicating that it may have neither a lower nor higher average emotional state. Similarly, the average emotional states of people with obsessive-compulsive disorder (OCD) have not been directly examined, however many indices of negative affect have been implicated in OCD \citep{ocdDisgust}. 
% In addition to the limited research available on average emotional states within each psychopathology, there has been no research examining potential differences in average emotional states across these MHCs and a control group. Emotion literature in regard to mental health % , psychological well-being, 
% and psychopathologies has predominately focused on emotion regulation issues through metrics of emotion reactivity and variability. 

% \subsubsection{Emotion Reactivity \& Psychopathology}
\noindent {\it Emotion Reactivity \& Psychopathology:}
Research has found that individuals with psychopathologies tend to take longer to recover from differing emotional states (i.e., emotional resetting or recovery rate) 
%and have greater emotional inertia 
than healthy individuals \citep{Kragel2022}. 
%These results were supported in a meta-analysis by \citet{Houben2015} where they found that emotional inertia was negatively related to indices of psychological well-being. 
That is, difficulty moving between emotional states is associated with lower psychological well-being. 
%The authors 
\citet{Houben2015}
also proposed that high emotional reactivity and slow recovery to baseline states is a maladaptive emotional pattern indicative of poor psychological well-being and psychopathology. In other words, people with poor psychological health may be highly reactive, emotionally volatile, and take a longer time to return to a baseline state.

\noindent {\it Emotion Variability \& Psychopathology:}
The \DT{\citet{Houben2015}} meta-analysis findings also indicate that higher emotional variability is related to lower psychological well-being. In particular, variability
%and instability were
was
%negatively
positively 
correlated with depression, anxiety, and other psychopathologies (e.g., bipolar, borderline personality disorder, etc.). This is supported by \citet{Heller2018} who found that higher positive and negative affect variability was associated with higher levels of depression and anxiety, these effects persisted for anxiety even after controlling for average positive affect. % and the rate of recovery to baseline (i.e., attractor strength). 
In contrast, variability was no longer associated with depression after controlling for average affect and the rate of recovery to baseline. This effect was attributed to anhedonia (the reduced ability to feel pleasure) which is a common symptom of depression that leads to reduced emotionality. 
% There has also been some preliminary evidence to suggest that emotional variability is associated with people with PTSD and panic disorder \citep{Pfaltz2010}. 

%In sum, 
\DT{Overall,}
%research on 
emotion dynamics
\DT{research}
suggests that one's average emotional state, emotional variability, rise rate, and recovery rate may vary by % one's 
their mental health. % (i.e., mentally ill or not). 
%Additionally, 
Preliminary research suggests that these metrics may also vary across different mental illnesses or psychopathologies. However, research on emotion dynamics within psychology and related fields has heavily relied on self-report measures and ecological momentary assessments (EMA). 
%However,
Affective self-report measures are subject to biases \citep{Kragel2022} and thus carry certain limitations (i.e., social pressures to be perceived as happy). Additionally, data collection with these methods is time-intensive thus, sample size and study length are limited. % Therewithal, 
Another window through which emotion dynamics can be inferred is through one's utterances \citep{movieED}.

\subsection{Utterance Emotion Dynamics}
%Research on emotion dynamics within psychology and related fields has heavily relied on self-report measures and ecological momentary assessments (EMA). However, affective self-report measures are subject to biases \citep{Kragel2022} and thus carry certain limitations (i.e., social pressures to be perceived as happy). 
%Further, data collection is time-intensive and the number of participants and duration over which measures can be recorded is limited.  
%Another window through which emotion dynamics can be inferred is through one's utterances \citep{movieED}. 
%Another window through which emotion dynamics can be inferred is through one's utterance \citep{movieED}.
%%%moved up to improve the transition

Early computational work on emotion change simply created emotion arcs using word--emotion association lexicons \cite{mohammad-2011-upon,emotionarcs}.
\DTtwo{\citet{teodorescu2023emoarc} proposed a mechanism to evaluate automatically generated emotion arcs and showed that lexicon-based methods obtain near-perfect correlations with the true emotion arcs.}
The Utterance Emotion Dynamics (UED) framework 
%measures patterns of emotion change from the emotion expressed in text through various metrics inspired by those in psychology. 
\DT{uses various metrics inspired by psychology research to quantify patterns of emotion change from the emotion expressed in text (from the emotion arcs).}
Using
\DT{a person's}
%one's 
utterances allows 
%us
\DT{researchers}
to analyse emotion dynamics 
\DT{since one's utterances can reasonably reflect one's thought process. UED allows for broader scale analyses}
%through various metrics on a broader scale 
across mediums (e.g., narratives, social media, etc.) and regions (e.g., cities, countries, etc.).
UED metrics have been used to study the emotional trajectories of movie characters \citep{movieED} and to analyse emotional patterns across geographic regions through Twitter data \citep{vishnubhotla-mohammad-2022-tusc}. 
\DT{\citet{seabrook} studied the association between depression severity and the emotion dynamics metric variability on Facebook and Twitter.}
The UED framework has also been applied to study developmental patterns of emotional change in poems written by children % across grades and 
%compared
% \DT{compared to UED of}
%to 
% adults 
\citep{teodorescu2023utterance}.

This work explores the relationship between UEDs and mental health conditions. Also, unlike past work in emotion dynamics that has focused on valence (pleasure--displeasure or positive--negative dimension), this work also explores the arousal (active--sluggish) and dominance (in control--out of control, powerful--weak) dimensions. Together, valence, arousal, and dominance are considered the core dimensions of emotion \cite{russell2003core}.

\section{Datasets}
\label{sec:datasets}
We use a recently compiled dataset---Twitter-STMHD \cite{_Singh_Arora_Shrivastava_Singh_Shah_Kumaraguru_2022}. 
% This large dataset 
It comprises of tweets from 27,003 users who have self-reported as having a mental health diagnosis on Twitter. The diagnoses % examined in this study 
include: depression, major depressive disorder (MDD), post-partum depression (PPD), post-traumatic stress disorder (PTSD), attention-deficit/hyperactivity disorders (ADHD), anxiety, bipolar, and obsessive-compulsive disorder (OCD).
\DT{We describe the dataset in Section \ref{sec:twitter_stmhd}, and our preprocessing steps in Section \ref{sec:preprocessing}.}

\DT{While our focus is on the relationship between emotion dynamics in tweets and MHCs, 
as a supplementary experiment, we also briefly explore the relationship between emotion dynamics in Reddit posts and depression.\footnote{The available Reddit data only included information about depression; we hope future work will explore other MHCs.} % as well.}
Textual datasets associated with MHCs are not common, but  
it is beneficial to contextualize findings on tweets in comparison to findings on datasets from other modalities of communication.
% Additionally, as a supplementary dataset we analysed data from Reddit (described in Section \ref{sec:eRisk_dataset}). While we focus on the relationship between UED and MHCs in tweets, it is beneficial to contextualize our findings on different datasets and modalities of communication. 
Due to the inherent differences in domains, dataset creation process, sample size, etc., we expect that there will be differences, however there may also be potential similarities in how UED relate to MHCs.} % and the control group. Therefore, the findings on this additional dataset adds to the generalizability of our results and claims.}

\subsection{Twitter Dataset: STMHD}
\label{sec:twitter_stmhd}
% To ensure a high-quality dataset, users were identified through several steps. 
\citet{_Singh_Arora_Shrivastava_Singh_Shah_Kumaraguru_2022} 
identified tweeters who self-disclosed an MHC diagnosis using carefully constructed regular expression patterns and manual verification. We summarize key details in the Appendix (Section \ref{app:twitter_dataset}). 
The control group consists of users identified from a random sample of tweets (posted during roughly the same time period as the MHC tweets). These tweeters did not post any tweets that satisfied the MHC regex described above.
% who are least likely to belong to a condition group 
% since they never self-disclosed any diagnoses. 
Additionally, users who had any posts about mental health discourse were removed from the control group. 
% We would note here 
Note that this does not guarantee that these users did not have an MHC diagnosis, but
%it is likely that
\DT{rather}
the set as a whole \DT{may have} 
%has
very few MHC tweeters.
The number of users in the control group was selected to match the size of the depression dataset, which has the largest number of users.

For the finalized set of users, four years of tweets were collected for each user: two years before self-reporting a mental health diagnosis and two years after. For the control group, tweets were randomly sampled from between January 2017 and May 2021 (same date range as other MHC classes).

% Users were filtered for \DT{their} number of tweets as well as their follower count.
Users with less than 50 tweets collected were removed so as to allow for more generalizable conclusions to be drawn. Similarly, users with more than 5000 followers were removed so as not to include celebrities, or other organizations that use Twitter 
%for discussing
\DT{to discuss}
well-being. 

\subsection{Reddit Dataset: eRisk 2018}
\label{sec:eRisk_dataset}
\DT{To further add to our findings, we also include the eRisk 2018 dataset \cite{eRisk2017, eRisk2018} in our experiments. % This dataset
It consists of users who self-disclosed as having depression on Reddit (expressions were manually checked), and a control group (individuals were randomly sampled). Contrary to the Twitter-STMHD dataset where users in the control group were removed for discussing well-being topics, \citet{eRisk2018} also considered users who discuss depression in the control group. % for a more realistic reflection of the population. 
The dataset includes several hundred posts per user, over approximately a 500-day period. We consider both the training set (which is from the eRisk 2017 task \cite{eRisk2017}) and the test set (from the eRisk 2018 task \cite{eRisk2018}).
}

\begin{table}[t]
\centering
{\small
\begin{tabular}
{llrr}
\hline
\multicolumn{1}{l}{\textbf{Dataset}} &
\multicolumn{1}{l}{\textbf{Group}} &
\multicolumn{1}{l}{\textbf{\#People}} & %Tweeters
\multicolumn{1}{l}{\textbf{Avg.\@ \#Posts/User}} \\
% \multicolumn{1}{l}{\textbf{Avg No. Words Per Tweet}}\\ 
\hline
\textit{Twitter} &  &  &  \\
& MHC & 10,069  & 2,177.4 \\
& $\;\;\;$ ADHD & 3,866 & 2,122.2  \\
& $\;\;\;$ Bipolar & 721 & 3,193.3  \\
& $\;\;\;$ Depression & 3,017 & 2,084.0   \\
& $\;\;\;$ MDD & 133 & 2,402.9 \\ 
& $\;\;\;$ OCD & 605 & 1,822.9  \\ 
& $\;\;\;$ PPD & 105 & 1,671.4  \\ 
& $\;\;\;$ PTSD & 1,622 & 1,944.9 \\
& Control & 4,097 & 1,613.6 \\
\textit{Reddit} &  &  &  \\
%& MHC &  &  \\
& Depression & 106 & 233.79 \\
& Control & 749 & 359.74 \\
\hline
\end{tabular}
}
 \vspace*{-1mm}
\caption{ The number of users in each mental health condition and 
%other details about 
the number of tweets per user in the preprocessed version of the Twitter-STMHD \DT{and Reddit eRisk datasets} we use for experiments.}
 \vspace*{-2mm}
\label{tab:twitter-stmhd_dataset_descriptives}
\end{table}

\subsection{Our Preprocessing}
\label{sec:preprocessing}

We further preprocessed 
\DT{both}
the Twitter-STMHD dataset 
\DT{and the eRisk dataset}
for our experiments (Section \ref{sec:experiments}), as
we are specifically interested in the unique patterns of UED for each disorder.
Several users self-reported as being diagnosed with more than one disorder, referred to as \textit{comorbidity}. 
We found a high comorbidity rate between users who self-reported as having anxiety and depression, as is also supported in the literature \citep{pollack2005comorbid,gorman,PMID:15014592,Cummings_2014}.
Therefore, we removed the anxiety class and only considered the depression class as it was 
\DT{the} %a
larger class between the two.
We also performed the following preprocessing steps: 

\begin{compactitem}
\item We only consider users who self-reported as having one disorder. We removed 1272 %512
users who had disclosed more than one diagnosis.
%(two disorders: 489, three disorders: 22, four disorders: 1).
\item  We only consider tweets in English, removing other languages.
\item We filtered out 
\DT{posts} %tweets
that contained URLs.
\item We removed retweets (identified through tweets containing `RT', `rt').
\item We computed the number of 
\DT{posts} %tweets
per user, and only considered users whose number of 
\DT{posts} %tweets
was within the interquartile range (between 25th and 75th percentile) for the diagnosis group they disclosed. This was to ensure that we are not including users that post very infrequently or very
% with very little content, or those who use social media extremely 
\DT{frequently}.
%regularly.  
\item \DT{We removed punctuation and stop words.} % from tweets.}
\end{compactitem}
 \noindent Table \ref{tab:twitter-stmhd_dataset_descriptives} shows key details of the filtered \DT{Twitter-STMHD} and Reddit e-risk datasets.
 \DT{We make our code for preprocessing the data publicly available.\footnote{\url{https://github.com/dteodore/EmotionArcs}}}

\section{Experiments}
\label{sec:experiments}

% \section{UED Metrics}
% \label{sec:ued_metrics}
%and shown their mathematical computation
% in the experiments described in Section \ref{sec:experiments}.

%\noindent \textbf{Emotional Home base}: 
%The baseline of one's average in a 2D emotional space (valence-arousal). 

\begin{figure}[t!]
    \centering
    \includegraphics[width=0.4\textwidth]
    %{images2/UED_Metrics_emnlp2023_new.png}
    {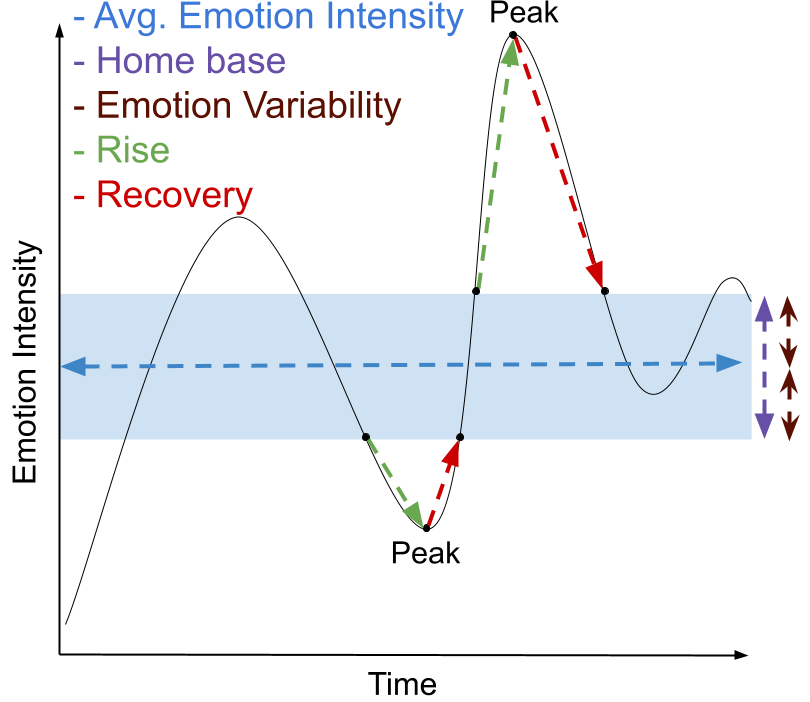}
% \vspace*{-5mm}
    \caption{
    Utterance emotion dynamics metrics quantify patterns of emotional change over time.
     }
     \label{fig:ued_metrics}
    % \vspace*{-2mm}
\end{figure}

To determine whether UED metrics from tweets can act as biosocial markers for psychopathology, we compare UED metrics for each MHC to the control group to see if they are statistically different. We compute the following UED metrics 
% (described in Section \ref{sec:ued_metrics}) 
per user in each condition, following the \textit{speaker UED} approach as described in \citet{teodorescu2023utterance}. 
\DT{In this approach, UED metrics are computed per speaker by placing all their utterances in temporal order and computing 
UED on these ordered utterances. 
\DTtwo{These metrics often rely on the characterization of the steady state or home base for emotions (regions of high probability).
\citet{movieED} define the region pertaining to one standard deviation on either side of the mean as the {\it home base}.}
Previous work computed speaker UED metrics for characters in movie dialogues \cite{movieED}, and for users on Twitter during the pandemic \cite{vishnubhotla-mohammad-2022-tusc}.}
We %briefly 
summarize \DT{the UED metrics} below (also see Figure \ref{fig:ued_metrics}):\\[-10pt]
%\begin{itemize}
\begin{compactitem}
\item \noindent \textbf{Average Emotional Intensity}: One's average emotion over time.\\[-10pt]
\item \noindent \textbf{Emotional Variability}: How much and how often one's emotional state changes over time.\\[-10pt]
\item \noindent \textbf{Rise Rate}: The rate at which one reaches peak emotional intensity from the home base (i.e., emotional reactivity).\\[-10pt]
\item \noindent \textbf{Recovery Rate}: The rate at which one recovers from peak emotional intensity to home base, (i.e., emotional regulation).\\[-10pt]
\end{compactitem}
%\end{itemize}
\noindent  % This approach is suitable since when we are interested in UED metrics at the aggregate level across groups, and therefore can compute measures per individual in each the group, to look at group differences.  
% \DTtwo{We depict each of these metrics in the Appendix (Section \ref{app:ued_metrics}).}

For each user, we order their tweets by timestamp and used the Emotion Dynamics toolkit \citep{VM2022-TED,movieED}\footnote{\url{https://github.com/Priya22/EmotionDynamics}} to compute UED metrics (average emotion, emotion variability, rise rate, and recovery rate).
We performed analyses for valence, arousal, and dominance.
For word-emotion association scores we use the NRC Valence, Arousal, and Dominance (VAD) lexicon \citep{vad-acl2018}.\footnote{\url{http://saifmohammad.com/WebPages/nrc-vad.html}}
It includes $\sim$20,000 common English words with scores ranging from $-1$ (lowest V/A/D) to $1$ (highest V/A/D). 
% and the NRC Emotion Intensity Lexicon (EIL) \citep{Mohammad13,mohammad-turney-2010-emotions} 
Afterwards, we performed an ANOVA to test for significant differences between groups in UED metrics, and post-hoc analyses to determine which groups specifically had significant differences from the control group.

\begin{table*}[t]
\centering
{\small
\begin{tabular}{lllllllllllllllll}
%{0.9\textwidth}{|*{13}{p{1cm}|}}
\hline
%\multicolumn{1}{l}{\textbf{\backslashbox{LowerText}{UpperText}}} &
&  & %\multicolumn{1}{l|}{\textbf{\diagbox[width=\dimexpr \textwidth/4+2\tabcolsep\relax,, height=0.7cm]{\\MHC--}{UED}}} &
% \multicolumn{3}{p{2.5cm}}{\centering \textbf{Average Emotion}} &
% \multicolumn{3}{p{2.5cm}}{\textbf{Emotion Variability}} &
% \multicolumn{3}{p{2.5cm}}{\textbf{Rise Rate}} &
% \multicolumn{3}{p{2.5cm}}{\textbf{Recovery Rate}}
  \multicolumn{3}{l}{\textbf{Average}} & & 
  \multicolumn{3}{l}{\textbf{Emotion}} & &
  \multicolumn{3}{c}{\textbf{Rise Rate}} & &
  \multicolumn{3}{l}{\textbf{Recovery}} \\ 
 & & %\multicolumn{1}{l|}{\textbf{Control}} & 
  \multicolumn{3}{l}{\textbf{Emotion}} & &
  \multicolumn{3}{l}{\textbf{Variability}} & &
  \multicolumn{3}{l}{\textbf{}}& &
  \multicolumn{3}{l}{\textbf{Rate}}\\
 \hline
\textbf{ Dataset} &\textbf{MHC--Control} & % &\multicolumn{1}{l|}{Emotion Dimension} & 
 \multicolumn{1}{l}{V} & \multicolumn{1}{l}{A} & \multicolumn{1}{l}{D} & & 
 \multicolumn{1}{l}{V} & \multicolumn{1}{l}{A} & \multicolumn{1}{l}{D} & &
 \multicolumn{1}{l}{V} & \multicolumn{1}{l}{A} & \multicolumn{1}{l}{D} & &
 \multicolumn{1}{l}{V} & \multicolumn{1}{l}{A} & \multicolumn{1}{l}{D} \\ \hline
 % & \multicolumn{1}{l|}{} & \multicolumn{1}{l|}{} &  & \multicolumn{1}{l|}{} & \multicolumn{1}{l|}{} &  & \multicolumn{1}{l|}{} & \multicolumn{1}{l|}{} &  & \multicolumn{1}{l|}{} & \multicolumn{1}{l|}{} &  \\
%& \multicolumn{1}{p{0.8cm}}{\textbf{V}} & & & & & \\
Twitter-STMHD & ADHD--control & $\downarrow$ & $\downarrow$ & $\downarrow$ &  &$\uparrow$ & $\uparrow$ & $\uparrow$  & & -- & -- & $\uparrow$ & & --  & $\uparrow$ & $\uparrow$\\ % (-.011, -.006)
& Bipolar--control & -- & $\downarrow$ & $\downarrow$ & & $\uparrow$ & $\uparrow$ & $\uparrow$ & & -- & -- & --   & &  $\uparrow$ & -- & --\\ %(-.018,-.011)

& MDD--control & $\downarrow$ & -- & $\downarrow$  && $\uparrow$ & $\uparrow$ & $\uparrow$ & & $\uparrow$ & -- & -- & & $\uparrow$ & $\uparrow$ & $\uparrow$\\ %(-.023,-.012)
& OCD--control & -- & $\downarrow$ & $\downarrow$ & &  $\uparrow$ & $\uparrow$ & $\uparrow$ & & -- & -- & $\uparrow$ & &  -- & $\uparrow$ & $\uparrow$\\ %(-.0136,-.007)
& PPD--control & -- & $\downarrow$ & $\downarrow$ & &  -- & $\uparrow$ & $\uparrow$ & &  -- & -- & -- & &  -- & -- & --\\ %(-.013,.002)
& PTSD--control & $\downarrow$ & -- & $\downarrow$ & & $\uparrow$ & $\uparrow$ & $\uparrow$ & & $\uparrow$ & $\uparrow$ & -- & & $\uparrow$ & $\uparrow$ & $\uparrow$ \\ %(-.018, -.013)

& Depression--control & -- & $\downarrow$ & $\downarrow$ & & $\uparrow$ & $\uparrow$ & $\uparrow$ & & $\uparrow$ & -- & $\uparrow$ & & $\uparrow$ & $\uparrow$ & $\uparrow$ \\ %(-.015,-.011)
Reddit eRisk& \DT{Depression--control} & -- & -- & $\downarrow$ & & $\uparrow$ & -- & $\uparrow$ & & -- & -- & -- & & -- & -- & -- \\

    \hline
%\end{tabulary} 
\end{tabular}
}
% \vspace*{-1mm}
\caption{ % \textbf{Valence (V), Arousal (A), Dominance (D)}: 
The difference in UED metrics % across MHC groups compared to 
between each MHC group and the control. A significant difference is indicated by an arrow; arrow direction indicates the direction of the difference. %; otherwise the cell has a dash. 
E.g., $\downarrow$ for ADHD--control and average emotion `V' %column 
means that the ADHD group has significantly lower average valence than the control group.
}
% \vspace*{-3mm}
\label{tab:arrows_all_emotions}
\end{table*}

\section{Results}
\label{sec:results}

To analyse potential differences across groups and the control group,
\DT{an ANOVA statistical test can be conducted, however several assumptions must be met. The three primary assumptions are: the data for each independent variable are approximately normally distributed, the data are independent of each other, and the distributions have roughly the same variance (homoscedasticity). We can assume the mean is \textit{normally distributed} according to the central limit theorem, due to the large sample size (law of large numbers). Since there are different tweeters in each MHC group we can assume the data are independent of each other. However, we note that people are largely not independent of each other e.g., having friends across groups, interacting with content from various mental health groups, etc. In our case, Levene’s test indicated that the assumption for homogeneity of variance was violated for all metrics and emotions (results in Appendix \ref{appendix:levenes}).
As such, we used Welch's ANOVA test which is an alternative to ANOVA when the equal variance assumption is not met.}
We conducted a 1-way Welch's ANOVA for each of the UED metrics (emotional average, emotional variability, rise-rate, and recovery rate) and emotions (valence, arousal, and dominance). We examined a total of \textit{N} = 14166 %14662 %15022 %15035 
users, see Table \ref{tab:twitter-stmhd_dataset_descriptives} for descriptives. 
% \DT{We report Welch's corrected results %below.
% in Table \ref{tab:welch}.
% }

% In order to conduct an ANOVA, several assumptions must be met. The three primary assumptions are: the data for each independent variable are approximately normally distributed, the data are independent of each other, and the distributions have roughly the same variance (homoscedasticity).
% We can assume our data is \textit{normally distributed} according to the central limit theorem, 
% %we assumed normality
% due to the large sample size. 
% Since there are different tweeters in each MHC group we can assume the data are independent of each other. However, we note that people are largely not independent of each other e.g., may have friends across groups, interact with content from various mental health groups, etc. 
% The \textit{homogeneity of variance} or homoscedasticity assumption can be tested by looking at the residuals and performing Levene's test. 
% In our case, %However,
% %our data did not meet this assumption since
% Levene’s test indicated that the assumption for homogeneity of variance was violated for all metrics and emotions (results in Appendix \ref{appendix:levenes}).
% As such, we used Welch's ANOVA test which is an alternative to ANOVA when the equal variance assumption is not met. 
% %"Welch's ANOVA", there's more than one Welch's Test 
% We report Welch's corrected results below. 
% %are reported. 

The first part of our analyses are Omnibus F-tests %, which test 
for significant differences between groups.
This test cannot tell us which groups are different from each other, just rather that there is a difference among groups.
For each combination of UED metrics and emotions (e.g., emotion variability for valence) there was a significant main effect which means we can conclude that at least one of the mental health diagnoses significantly differed. 
%(except arousal rise rate).
%for each of the UED metrics across all three emotions.
\DTtwo{We show the degrees of freedom, F-statistic, p-value, and corrected effect size in Table \ref{tab:welchs_valence_arousal_dominance} in the Appendix for valence, arousal, and dominance.}
% for valence. Results for arousal and dominance are shown in Table \ref{tab:welchs_arousal_dominance} in the Appendix.
The effect size tells us how meaningful the difference between groups is for each metric.\footnote{An effect size <0.01 is \textit{very small}, between 0.01 to 0.06 is \textit{small}, between 0.06 to 0.14 is \textit{medium}, and greater than or equal to 0.14 is \textit{large}.} For example, 
%3.23\%
3.31\%
of the total variance in the emotional variability for valence is accounted for by diagnosis (small effect).
%include the statistic in brackets (the effect size) 

% The results
% indicated a significant main effect of emotional average 
% (\textit{F}(7, 1206.21)=15.07, \textit{p}<.001,
% \textit{est $\omega$}\textsuperscript{2}=), 
% emotional variability
% (\textit{F}(7, 1206.23)=73.19, \textit{p}<.001,
% \textit{est $\omega$}\textsuperscript{2}= ),
% rise rate
% (\textit{F}(7, 1211.29)=11.22, \textit{p}<.001, \textit{est $\omega$}\textsuperscript{2}= ),
% and recovery rate
% (\textit{F}(7, 1211.67)=11.60, \textit{p}<.001, \textit{est $\omega$}\textsuperscript{2}= ). 
%Thus, we can conclude that at least one of the mental health diagnoses significantly differed for each of the UED metrics.
%Since assumptions for homogeneity of variance were violated, a Games-Howell correction was applied to posthoc analyses (See Table \ref{tab:adult_ued}). 
%Add table for all comparisons
%can you get the lowercase omega symbol to work?

% In the next step of our analyses, 
Next, we would like to know exactly which groups differed from the control group. In order to do this, we performed post hoc analyses for pairwise comparisons between groups for each metric across the three \DT{dimensions} of emotions. We applied a Games-Howell correction since the assumption for homogeneity of variance was violated. In the following Sections we detail how the UED metrics compare across MHCs compared to the control group.
In Table \ref{tab:arrows_all_emotions} we show the \DT{pairwise comparison} results for which UED metrics and emotion combination significantly differed from the control group across diagnoses, and the direction of the difference.
% We also report the means for each UED metric and group in Appendix \ref{app:avg_ued} for valence, arousal and dominance (Table \ref{tab:avg_valence}, \ref{tab:avg_arousal} and \ref{tab:avg_dominance}).
\DT{
%Posts from users in both groups where collected (up to 1000 posts + 1000 comments). 
 We also show results on the eRisk dataset in Table \ref{tab:arrows_all_emotions}, and contrast our findings for depression between the Twitter and Reddit datasets. 
}

We contextualize our results with previous findings in psychology and studies in NLP. We note that the relationship between patterns of emotion change and well-being for the dimensional emotions arousal and dominance are under-explored -- our findings provide important benchmarks for these emotions and UED metrics more generally.

\subsection{How does the average emotion %across 
for an MHC
%a mental health condition 
compare to the control?}

\noindent \textbf{Valence:} The average valence was significantly lower for the ADHD, 
MDD,
%depression,
and PTSD groups compared to the control group.

\noindent \textbf{Arousal:}
The ADHD, depression, bipolar, PPD, and OCD groups showed significantly lower arousal compared to the control group.

\noindent \textbf{Dominance:}
All MHC groups (ADHD, depression, bipolar, MDD, PPD, PTSD, OCD) showed significantly lower dominance compared to the control group. 
\DT{Additionally, the depression group in the eRisk dataset also had significantly lower dominance compared to the control group.}

\noindent \textit{\textbf{Discussion}:} Our findings align with results in psychology, and NLP studies looking at the average emotion \DT{intensity} expressed in text. Valence was found to be lower in individuals with depression (of which MDD is a type of depression) through self-reports questionnaires 
 \cite{Heller2018,depressedYouth} and on social media \citep{seabrook,DeChoudhury,De_Choudhury_Gamon_Counts_Horvitz_2021}. Further, work in psychology has found individuals with PTSD and ADHD have lower valence \citep{Pugach2023, STICKLEY2018317}. It has also been shown that lower arousal and dominance in speech is associated with depression \citep{StasakEppsCumminsetal2016,Osatuke,Gumus}. While average emotion \DT{intensity} is one of the more commonly explored measures, % of emotion,
 there is still relatively few works studying the relationships between arousal and dominance with mental health, compared to valence. Interestingly, dominance appears to differ for many MHCs (all studied here) from the control group, pointing towards an important indicator of well-being.

\subsection{How does emotion variability % across 
for an MHC %a mental health condition 
compare to the control?}

\noindent \textbf{Valence:} Variability for valence was significantly higher for the ADHD, depression, bipolar, MDD, PTSD, and OCD groups compared to the control. PPD did not show differences from the control.
\DT{The depression group in the eRisk dataset also showed significantly higher valence variability compared to the control group.}

\noindent \textbf{Arousal:}
All MHC groups (ADHD, depression, bipolar, MDD, PPD, PTSD, and OCD) showed significantly higher arousal variability.

\noindent \textbf{Dominance:}
All MHC groups (ADHD, depression, bipolar, MDD, PPD, PTSD, OCD) had significantly higher dominance variability than the control group.
\DT{The depression group in the eRisk dataset also had significantly higher dominance variability compared to the control group.}

\noindent \textit{\textbf{Discussion}:} In several studies in psychology, it has been shown that higher valence variability occurred for individuals with depression, PTSD \citep{Houben2015, Heller2018} and is negatively correlated with overall well-being \citep{Houben2015}.
%Similarly, \citet{seabrook} found higher valence variability on Facebook indicated higher depression severity. Interestingly, this contradicted their findings on Facebook. 
Interestingly, \citet{seabrook} found higher valence variability on Twitter indicated lower depression severity which contradicted their findings on Facebook. 
\citet{Kuppens2007} report that valence variability was negatively related to self-esteem and was positively related to neuroticism and depression.
Overall, our results align with emotional variability having strong ties with well-being. Arousal and dominance variability appear to be \textit{biosocial} markers across several MHCs, although minimally explored in the literature (\citet{RANNEY2020101542} found higher affective arousal variability was associated with generalized anxiety disorder).

\subsection{How does emotional rise rate %across 
for an MHC
% a mental health condition 
compare to the control?}

\noindent \textbf{Valence:} Rise-rate for valence was significantly higher for the depression, MDD, and PTSD groups compared to the control group. 

\noindent \textbf{Arousal:} PTSD was the only group which had statistically higher arousal rise rates than the control group.
%None of the MHC groups has statistically different arousal rise-rates than the control group. 

\noindent \textbf{Dominance:}
The ADHD, depression, and OCD groups had significantly higher rise rates than the control group.

\noindent \textit{\textbf{Discussion}:}
Rise-rate is analogous to emotional reactivity in psychology, and quickly moving to \DT{peak} emotional states has been shown in individuals with maladaptive emotion patterns and lower psychological well-being \cite{Houben2015}. It is interesting to note that valence and dominance rise rates differed across MHC to the control, whereas not to the same extent for arousal.

\subsection{How does emotional recovery rate %across 
for an MHC
% a mental health condition 
compare to the control?}

\noindent \textbf{Valence:} Recovery rate for valence was significantly higher for the %ADHD, Bipolar disorder
depression, bipolar,
%disorder,
MDD, and PTSD groups compared to the control group. 

\noindent \textbf{Arousal:}
The ADHD, depression,
MDD, %new
PTSD, and OCD groups showed significantly higher arousal recovery rates % compared to 
than the control group.

\noindent \textbf{Dominance:}
The ADHD, depression, MDD, PTSD and OCD groups showed significantly higher \DT{dominance} recovery rates than the control group.

\noindent \textit{\textbf{Discussion}:}
Recovery rates can be thought of as a proxy of emotion regulation, and slower recovery from emotional events is associated with psychopathology and poor psychological well-being \cite{Houben2015,Kragel2022}. Our results, while pointing to higher recovery rates, indicate significant differences from the control group. 
This is an interesting result that can be further explored if found in other mediums such as Reddit.
%This could be due to the brevity of tweets, or other aspects such as how stages of emotional regulation may not be expressed in tweets.
%users mainly express themselves for reporting emotionally charged events and do not post their feelings during recovery periods.   

\DT{
We expected to see differences in which UED significantly different from the control between the Twitter and Reddit datasets due to the different mediums, collection process, etc.\@ as mentioned in Section \ref{sec:datasets}. However, we also saw commonalities in that average dominance was significantly lower, and valence and dominance variability were significantly higher for the depression group compared to the control. It appears that despite dataset differences, these \textit{biosocial} markers are strong indicators for depression. Future work can explore further whether rise and recovery rates act as indicators as well, perhaps by varying parameters involved in computing UED, such as the time-step granularity. % at which emotion dynamics are computed. 

Overall, we notice that there are quite a few UED metrics across emotion dimensions which are significantly different for MHCs from the control group. Often also, that the direction of difference is similar for valence, arousal, and dominance. Work in psychology supports that often valence and dominance move in similar directions. For example, \cite{feelings2015} found software engineers were more productive when in a good mood associated with higher valence (i.e., more positive) and higher dominance (i.e., more in control of one's feelings).
%Whereas less productive when in a more negative state with lower valence and lower dominance.
Likewise, 
There may be several factors which influence valence and dominance to move in similar directions in the control group and MHCs.
% and likewise some other factors which influence valence and dominance to move in opposite directions. 
% Future work should investigate to what extent emotion dimensions correlate ...
}

\begin{figure*}[h]
    \centering
    \begin{subfigure}[b]{0.45\textwidth}
    \centering
    \subcaption{}
    \includegraphics[width=\textwidth]{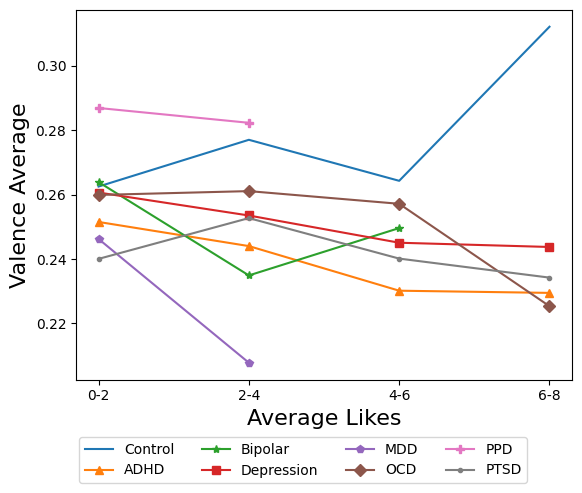}
    % \caption{}
    \label{fig:emo_mean_val}
    \end{subfigure}
    \hfill
    \begin{subfigure}[b]{0.45\textwidth}
    \centering
    \subcaption{}
    \includegraphics[width=\textwidth]{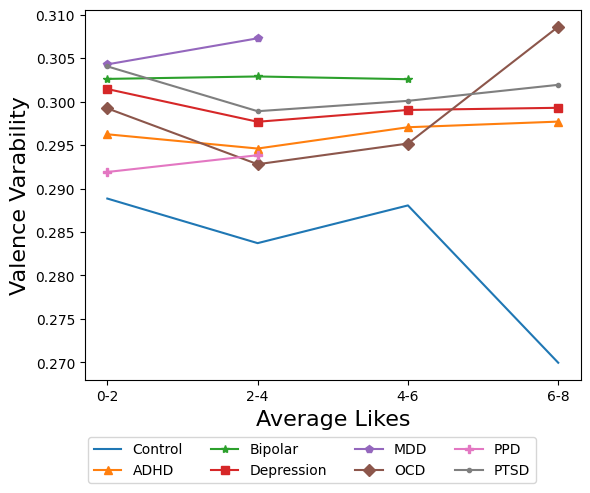}
    % \caption{}
    \label{fig:emo_std_val}
    \end{subfigure}
    % \vskip\baselineskip
    % \vspace*{-8mm}
    % \centering
    % \begin{subfigure}[b]{0.45\textwidth}
    % \centering
    % \subcaption{}
    % \includegraphics[width=\textwidth]{images2/Rise_Rate_Valence_avg_likes_new_chart_Oct22.png}
    % % \caption{}
    % \label{fig:rise_rate_val}
    % \end{subfigure}
    % \hfill
    % \begin{subfigure}[b]{0.45\textwidth}
    % \centering
    % \subcaption{}
    % \includegraphics[width=\textwidth]{images2/Recovery_Rate_Valence_avg_likes_new_chart_Oct22.png}
    % % \caption{}
    % \label{fig:recovery_rate_val}
    % \end{subfigure}
    \vspace*{-3mm}
    \caption{% UED metrics for 
    \textbf{Valence} average and variability across %various 
    levels of user \textit{popularity} on Twitter (average \#likes on posts). % received per post.
    }
    \label{fig:ued_val_avg_likes}
\end{figure*}

\section{Results: \textit{Biosocial} Aspects}

We now explore whether the UED metrics that were found to be significant indicators of certain MHCs, continue to remain discriminating even when accounting for certain social factors such as how popular a user's posts tend to be.
Metadata for 
\DT{the number of likes per post is available in the dataset.}
% this
% %these
% \textit{popularity} measure is available for each user, 
\DT{So, we simply 
extracted 
this information and 
%the number of followers, 
%the number of people following,
% and 
computed the average number of likes per tweeter.}
%it 
% \DT{Looking at the distribution for 
% users' average number of likes,
% %each of these social aspects,
% we noticed the range most users were in, and selected the number of bins and a bin width to analyse UED across various levels of likes.
% %\textit{popularity}. 
% % For the average number of likes received 
\DT{Most users had 0--4 likes per post in both the MHCs and the control group, up to an average of 8 likes per post.} 
% Therefore, we determined the optimal bin width is two, and looked at four bins in total.}
\DT{ %Afterwards, 
We compare the average UED per emotion dimension across the various bins of 
average likes.
%the popularity measure. 
If a popularity measure 
% such as this
does not impact UED, then we would expect to see consistent differences between the control group and an MHC across popularity bins.
However, if a popularity measure does influence % why we saw differences in 
UED then the differences across MHCs and the control may disappear when controlling for popularity.}
% between an MHC and the control group, then we should be varying differences in UED across popularity bins. }

\DT{In Figure \ref{fig:ued_val_avg_likes}, we show the 
two
%four
UED metrics (e.g., average emotion intensity and emotional variability) 
%rise rate, and recovery rate) 
for valence, across various bins of average likes per tweeter for each group. 
In the Appendix (Section 
\ref{app:ued_likes_valence},
\ref{app:ued_likes_arousal}, and \ref{app:ued_likes_dominance}), we show the \DTtwo{remaining} results for arousal and dominance.
If a bin 
%of the popularity aspect
had less than ten users in it (e.g., average of 6--8 likes per post), then we did not plot a value for it.
In each figure the control group is the blue line (no shapes).
\DTtwo{For MHCs which significantly differed from the control for an emotion and UED pair, we show the mean difference 
% between groups
in Appendix \ref{app:mean_diff}. If this difference between an MHC and the control remains fairly consistent across ranges of average likes, this UED is most likely not influenced by this social aspect.}\\[3pt]
%Avg
\noindent \textit{Average Valence:} In the previous section, we showed that average valence is markedly lower for ADHD, MDD, and PTSD. From Figure \ref{fig:ued_val_avg_likes} (a), we observe that the average valence for these MHCs is markedly lower than control for various Average-Likes bins. (We also observe markedly greater differences when average likes is in the [6,8) range.)\\[3pt] 
\noindent \textit{Valence Variability:} In the previous section, we showed that valence variability is markedly higher for all MHCs except PPD. From Figure \ref{fig:ued_val_avg_likes} (b), we observe that the valence variability for these MHCs is markedly higher than control for various Average-Likes bins. (We also observe markedly lower variability for control when average likes is in the [6,8) range.)\\[3pt] 
\noindent \textit{Valence Rise Rate:} In the previous section, we showed that valence rise rate is markedly higher for MDD, PTSD, and Depression. From Figure \ref{fig:ued_val_avg_likes2} \DTtwo{(a)}
%(c),
we observe that the valence rise rate for these MHCs is markedly higher than control for various Average-Likes bins. (We also observe markedly lower valence rise rate for control when average user likes is in the [6,8) range.)\\[3pt] 
\noindent \textit{Valence Recovery Rate:} In the previous section, we showed that valence recovery rate is markedly higher for Bipolar, MDD, PTSD, and Depression. From Figure \ref{fig:ued_val_avg_likes2} \DTtwo{(b)}
% (d),
we observe that the valence recovery rate for these MHCs is markedly higher than control for various Average-Likes bins. (We also observe markedly lower 
%valence
recovery rate for control when average user likes is in the [6,8) range.) \\[6pt] 
\indent Thus, overall, we observe that for the 
valence UEDs which showed significant differences across MHCs and control, these differences do not disappear when accounting for popularity measures such as the average number of likes a user's posts get. Similar trends overall were found for arousal and dominance 
% (shown in the Appendix).
(Appendix \ref{app:ued_likes_arousal} and \ref{app:ued_likes_dominance}).
UED
for all three emotion \DT{dimensions} 
% (for valence and dominance)
appear to be robust indicators for various mental health conditions, even when
\DT{users have some varying social characteristics.}
}

\section{Conclusion}
We showed for the first time that there are significant relationships between patterns of emotion change in text written by individuals with a self-disclosed MHC compared to a control group. % By using a Twitter dataset where users have chosen to disclose an MHC diagnosis, 
Specifically, we found significant differences in four utterance emotion dynamics (UED) metrics (average emotion, emotion variability, rise rate, and recovery rate) across three emotion \DT{dimensions} (valence, arousal, and dominance) for 7 MHCs. Our findings provide important contextual information of overall well-being and supporting indicators (in addition to other assessments) to clinicians for diagnosis detection and management. 

%Building on this work, we plan to extend our analyses on more extensive data from Reddit. %and % consider % other mediums such as other social media platforms (e.g., Reddit), and 
Looking ahead, we plan to explore UED metrics %are biosocial markers 
across different textual genres, regions, languages, and demographics (such as socioeconomic status), in collaboration with clinicians.
% . Notably, we will collaborate with clinicians to gains a more in-depth understanding of various aspects such as recovery and intervention, and the role emotion dynamics may play. 
Through such a partnership, UED metrics could be studied in the context of clinical information as well. 
Lastly, % we are developing 
exploring new UED metrics 
% which have been shown in psychology to 
that correlate with psychopathology is a promising direction for future work.

% and plan to assess whether these metrics may also be biosocial markers, and which may be more suitable based on emotion, domain, etc.       

\section*{Limitations}
In this study, we used NLP techniques to compare 
%the
UED across different psychopathologies and a control group. It is important to be cautious of interpreting these results due to natural limitations within the dataset. Due to the high rates of comorbidity with mental health disorders \citep{Druss} examining users who only disclosed one diagnosis may not be representative of the population. Furthermore, it is also possible that the dataset included users with more than one disorder but only disclosed one (e.g., a user may have been diagnosed with depression and ADHD but only tweeted "diagnosed with ADHD" or vice versa). Self-disclosure of diagnoses may also be inaccurate due to reasons such as impression management \citep{LEARY2001} or social desirability \citep{LATKIN2017} where users may disclose having a diagnosis without having received a formal diagnosis. Alternatively, there may have been users included in the control group who have a formal diagnosis of one or more mental health disorders but did not disclose 
this
%their mental disorders
on Twitter. 
Overall,
%In sum, 
despite the
dataset creators'
%authors' 
best efforts to collect data accordingly, the users in each group may not be representative of the mental disorders. 
Future research
could %should
replicate this study using a sample of users with confirmed formal diagnoses. 

%EMNLP 2023 requires all submissions to have a section titled ``Limitations'', for discussing the limitations of the paper as a complement to the discussion of strengths in the main text. This section should occur after the conclusion, but before the references. It will not count towards the page limit.  

%The discussion of limitations is mandatory. Papers without a limitation section will be desk-rejected without review.
%ARR-reviewed papers that did not include ``Limitations'' section in their prior submission, should submit a PDF with such a section together with their EMNLP 2023 submission.

%While we are open to different types of limitations, just mentioning that a set of results have been shown for English only probably does not reflect what we expect. 
%Mentioning that the method works mostly for languages with limited morphology, like English, is a much better alternative.
%In addition, limitations such as low scalability to long text, the requirement of large GPU resources, or other things that inspire crucial further investigation are welcome.

\section*{Ethics Statement}
\noindent Our research interest is to study emotions at an aggregate/group level. This has applications in emotional development psychology and in public health (e.g., overall well-being and mental health). However, emotions are complex, private, and central to an individual's experience. Additionally, each individual expresses emotion differently through language, which results in large amounts of variation. Therefore, several ethical considerations should be accounted for when performing any textual analysis of emotions \cite{Mohammad22AER,mohammad2020practicaleacl}.
The ones we would particularly like to highlight are listed below:
\begin{compactitem}
    \item Our work on studying emotion word usage should not be construed as detecting how people feel; rather, we draw inferences on the emotions that are conveyed by users via the language that they use. 
    \item The language used in an utterance may convey information about the emotional state (or perceived emotional state) of the speaker, listener, or someone mentioned in the utterance. However, it is not sufficient for accurately determining any of their momentary emotional states. Deciphering the true momentary emotional state of an individual requires extra-linguistic context and world knowledge.
    Even then, one can be easily mistaken.
    \item The inferences we draw in this paper are based on aggregate trends across large populations. We do not draw conclusions about specific individuals or momentary emotional states.
\end{compactitem}
% Scientific work published at EMNLP 2023 must comply with the \href{https://www.aclweb.org/portal/content/acl-code-ethics}{ACL Ethics Policy}. We encourage all authors to include an explicit ethics statement on the broader impact of the work, or other ethical considerations after the conclusion but before the references. The ethics statement will not count toward the page limit (8 pages for long, 4 pages for short papers).

\section*{Acknowledgements}
Many thanks to Krishnapriya Vishnubhotla for the Emotion Dynamics codebase, which set the groundwork for computing emotion dynamics, and for insightful discussions.
This research was supported by NSERC, SSHRC, Digital Research Alliance of Canada (alliancecan.ca), Alberta Innovates, DeepMind, and CIFAR. Alona Fyshe holds a Canada CIFAR AI Chair.
This research project is funded by the Bavarian Research Institute for Digital Transformation (bidt), an institute of the Bavarian Academy of Sciences and Humanities. The author is responsible for the content of this publication.

% Entries for the entire Anthology, followed by custom entries
\bibliography{anthology,custom}

\begin{thebibliography}{52}
\expandafter\ifx\csname natexlab\endcsname\relax\def\natexlab#1{#1}\fi

\bibitem[{Ballman(2015)}]{biomarker}
{Karla V.} Ballman. 2015.
\newblock \href {https://doi.org/10.1200/JCO.2015.63.3651} {Biomarker: Predictive or prognostic?}
\newblock \emph{Journal of Clinical Oncology}, 33(33):3968--3971.
\newblock Publisher Copyright: {\textcopyright} 2015 American Society of Clinical Oncology. All rights reserved.

\bibitem[{Calzà et~al.(2021)Calzà, Gagliardi, {Rossini Favretti}, and Tamburini}]{CALZA2021101113}
Laura Calzà, Gloria Gagliardi, Rema {Rossini Favretti}, and Fabio Tamburini. 2021.
\newblock \href {https://doi.org/https://doi.org/10.1016/j.csl.2020.101113} {Linguistic features and automatic classifiers for identifying mild cognitive impairment and dementia}.
\newblock \emph{Computer Speech \& Language}, 65:101113.

\bibitem[{Corcoran et~al.(2020)Corcoran, Mittal, Bearden, {E. Gur}, Hitczenko, Bilgrami, Savic, Cecchi, and Wolff}]{CORCORAN2020158}
Cheryl~M. Corcoran, Vijay~A. Mittal, Carrie~E. Bearden, Raquel {E. Gur}, Kasia Hitczenko, Zarina Bilgrami, Aleksandar Savic, Guillermo~A. Cecchi, and Phillip Wolff. 2020.
\newblock \href {https://doi.org/https://doi.org/10.1016/j.schres.2020.04.032} {Language as a biomarker for psychosis: A natural language processing approach}.
\newblock \emph{Schizophrenia Research}, 226:158--166.
\newblock Biomarkers in the Attenuated Psychosis Syndrome.

\bibitem[{Cummings et~al.(2014)Cummings, Caporino, and Kendall}]{Cummings_2014}
Colleen~M. Cummings, Nicole~E. Caporino, and Philip~C. Kendall. 2014.
\newblock \href {https://doi.org/10.1037/a0034733} {Comorbidity of anxiety and depression in children and adolescents: 20 years after.}
\newblock \emph{Psychological Bulletin}, 140(3):816--845.

\bibitem[{Cuteri et~al.(2022)Cuteri, Minori, Gagliardi, Tamburini, Malaspina, Gualandi, Rossi, Moscano, Francia, and Parmeggiani}]{Cuteri}
Vittoria Cuteri, Giulia Minori, Gloria Gagliardi, Fabio Tamburini, Elisabetta Malaspina, Paola Gualandi, Francesca Rossi, Milena Moscano, Valentina Francia, and Antonia Parmeggiani. 2022.
\newblock \href {https://doi.org/10.1007/s40519-021-01273-7} {Linguistic feature of anorexia nervosa: a prospective case-control pilot study}.
\newblock \emph{Eating and weight disorders : EWD}, 27(4):1367—1375.

\bibitem[{Davidson(1998)}]{davidson1998}
Richard~J. Davidson. 1998.
\newblock \href {https://doi.org/10.1080/026999398379628} {Affective style and affective disorders: Perspectives from affective neuroscience}.
\newblock \emph{Cognition and Emotion}, 12(3):307--330.

\bibitem[{De~Choudhury et~al.(2013)De~Choudhury, Counts, and Horvitz}]{DeChoudhury}
Munmun De~Choudhury, Scott Counts, and Eric Horvitz. 2013.
\newblock \href {https://doi.org/10.1145/2464464.2464480} {Social media as a measurement tool of depression in populations}.
\newblock In \emph{Proceedings of the 5th Annual ACM Web Science Conference}, WebSci '13, page 47–56, New York, NY, USA. Association for Computing Machinery.

\bibitem[{De~Choudhury et~al.(2021)De~Choudhury, Gamon, Counts, and Horvitz}]{De_Choudhury_Gamon_Counts_Horvitz_2021}
Munmun De~Choudhury, Michael Gamon, Scott Counts, and Eric Horvitz. 2021.
\newblock \href {https://doi.org/10.1609/icwsm.v7i1.14432} {Predicting depression via social media}.
\newblock \emph{Proceedings of the International AAAI Conference on Web and Social Media}, 7(1):128--137.

\bibitem[{Druss and Walker(2011)}]{Druss}
Benjamin~G Druss and Elizabeth~Reisinger Walker. 2011.
\newblock \href {http://europepmc.org/abstract/MED/21675009} {Mental disorders and medical comorbidity}.
\newblock \emph{The Synthesis project. Research synthesis report}, (21):1—26.

\bibitem[{Gagliardi and Tamburini(2021)}]{Gloria}
Gloria Gagliardi and Fabio Tamburini. 2021.
\newblock \href {https://doi.org/10.1418/101111} {Linguistic biomarkers for the detection of mild cognitive impairment}.
\newblock \emph{Lingue e linguaggio, Rivista semestrale}, (1/2021):3--31.

\bibitem[{Gagliardi and Tamburini(2022)}]{gagliardi-tamburini-2022-automatic}
Gloria Gagliardi and Fabio Tamburini. 2022.
\newblock \href {https://aclanthology.org/2022.lrec-1.561} {The automatic extraction of linguistic biomarkers as a viable solution for the early diagnosis of mental disorders}.
\newblock In \emph{Proceedings of the Thirteenth Language Resources and Evaluation Conference}, pages 5234--5242, Marseille, France. European Language Resources Association.

\bibitem[{Gkotsis et~al.(2016)Gkotsis, Oellrich, Hubbard, Dobson, Liakata, Velupillai, and Dutta}]{gkotsis-etal-2016-language}
George Gkotsis, Anika Oellrich, Tim Hubbard, Richard Dobson, Maria Liakata, Sumithra Velupillai, and Rina Dutta. 2016.
\newblock \href {https://doi.org/10.18653/v1/W16-0307} {The language of mental health problems in social media}.
\newblock In \emph{Proceedings of the Third Workshop on Computational Linguistics and Clinical Psychology}, pages 63--73, San Diego, CA, USA. Association for Computational Linguistics.

\bibitem[{Gorman(1996)}]{gorman}
Jack~M. Gorman. 1996.
\newblock \href {https://doi.org/https://doi.org/10.1002/(SICI)1520-6394(1996)4:4<160::AID-DA2>3.0.CO;2-J} {Comorbid depression and anxiety spectrum disorders}.
\newblock \emph{Depression and Anxiety}, 4(4):160--168.

\bibitem[{Graziano et~al.(2007)Graziano, Reavis, Keane, and Calkins}]{success}
Paulo~A. Graziano, Rachael~D. Reavis, Susan~P. Keane, and Susan~D. Calkins. 2007.
\newblock \href {https://doi.org/https://doi.org/10.1016/j.jsp.2006.09.002} {The role of emotion regulation in children's early academic success}.
\newblock \emph{Journal of School Psychology}, 45(1):3--19.

\bibitem[{Graziotin et~al.(2015)Graziotin, Wang, and Abrahamsson}]{feelings2015}
Daniel Graziotin, Xiaofeng Wang, and Pekka Abrahamsson. 2015.
\newblock \href {https://doi.org/10.1002/smr.1673} {Do feelings matter? on the correlation of affects and the self-assessed productivity in software engineering}.
\newblock \emph{J. Softw. Evol. Process}, 27(7):467–487.

\bibitem[{Gumus et~al.(2023)Gumus, DeSouza, Xu, Fidalgo, Simpson, and Robin}]{Gumus}
Melisa Gumus, Danielle~D DeSouza, Mengdan Xu, Celia Fidalgo, William Simpson, and Jessica Robin. 2023.
\newblock \href {https://doi.org/10.1177/20552076231180523} {Evaluating the utility of daily speech assessments for monitoring depression symptoms}.
\newblock \emph{DIGITAL HEALTH}, 9:20552076231180523.

\bibitem[{Heller et~al.(2018)Heller, Fox, and Davidson}]{Heller2018}
Aaron~S Heller, Andrew~S Fox, and Richard~J Davidson. 2018.
\newblock Parsing affective dynamics to identify risk for mood and anxiety disorders.
\newblock \emph{Emotion}, 19(2):283--291.

\bibitem[{Hipson and Mohammad(2021)}]{movieED}
Will~E. Hipson and Saif~M. Mohammad. 2021.
\newblock \href {https://doi.org/10.1371/journal.pone.0256153} {Emotion dynamics in movie dialogues}.
\newblock \emph{PLOS ONE}, 16(9):1--19.

\bibitem[{Hirschfeld(2001)}]{PMID:15014592}
Robert M.~A. Hirschfeld. 2001.
\newblock \href {https://doi.org/10.4088/pcc.v03n0609} {The comorbidity of major depression and anxiety disorders: Recognition and management in primary care}.
\newblock \emph{Primary care companion to the Journal of clinical psychiatry}, 3(6):244—254.

\bibitem[{Houben et~al.(2015)Houben, Van Den~Noortgate, and Kuppens}]{Houben2015}
Marlies Houben, Wim Van Den~Noortgate, and Peter Kuppens. 2015.
\newblock The relation between short-term emotion dynamics and psychological well-being: A meta-analysis.

\bibitem[{Koops et~al.(2023)Koops, Brederoo, de~Boer, Nadema, Voppel, and Sommer}]{koops}
Sanne Koops, Sanne~G Brederoo, Janna~N de~Boer, Femke~G Nadema, Alban~E Voppel, and Iris~E Sommer. 2023.
\newblock \href {https://doi.org/10.2174/1871527320666211213125847} {Speech as a biomarker for depression}.
\newblock \emph{CNS\&; neurological disorders drug targets}, 22(2):152—160.

\bibitem[{Kragel et~al.(2022)Kragel, Hariri, and LaBar}]{Kragel2022}
Philip~A. Kragel, Ahmad~R. Hariri, and Kevin~S. LaBar. 2022.
\newblock \href {https://doi.org/10.1162/jocn_a_01787} {The temporal dynamics of spontaneous emotional brain states and their implications for mental health}.
\newblock \emph{Journal of cognitive neuroscience}, 34(5):715--728.
\newblock May, 2022.

\bibitem[{Kuppens et~al.(2007)Kuppens, Van~Mechelen, Nezlek, Dossche, and Timmermans}]{Kuppens2007}
Peter Kuppens, Iven Van~Mechelen, John~B Nezlek, Dorien Dossche, and Tinneke Timmermans. 2007.
\newblock \href {https://doi.org/10.1037/1528-3542.7.2.262} {Individual differences in core affect variability and their relationship to personality and psychological adjustment}.
\newblock \emph{Emotion (Washington, D.C.)}, 7(2):262—274.

\bibitem[{Kuppens and Verduyn(2015)}]{ed}
Peter Kuppens and Philippe Verduyn. 2015.
\newblock \href {https://doi.org/10.1080/1047840X.2015.960505} {Looking at emotion regulation through the window of emotion dynamics}.
\newblock \emph{Psychological Inquiry}, 26(1):72--79.

\bibitem[{Kuppens and Verduyn(2017)}]{KUPPENS201722}
Peter Kuppens and Philippe Verduyn. 2017.
\newblock \href {https://doi.org/https://doi.org/10.1016/j.copsyc.2017.06.004} {Emotion dynamics}.
\newblock \emph{Current Opinion in Psychology}, 17:22--26.
\newblock Emotion.

\bibitem[{Latkin et~al.(2017)Latkin, Edwards, Davey-Rothwell, and Tobin}]{LATKIN2017}
Carl~A. Latkin, Catie Edwards, Melissa~A. Davey-Rothwell, and Karin~E. Tobin. 2017.
\newblock \href {https://doi.org/https://doi.org/10.1016/j.addbeh.2017.05.005} {The relationship between social desirability bias and self-reports of health, substance use, and social network factors among urban substance users in baltimore, maryland}.
\newblock \emph{Addictive Behaviors}, 73:133--136.

\bibitem[{Leary(2001)}]{LEARY2001}
M.R. Leary. 2001.
\newblock \href {https://doi.org/https://doi.org/10.1016/B0-08-043076-7/01727-7} {Impression management, psychology of}.
\newblock In Neil~J. Smelser and Paul~B. Baltes, editors, \emph{International Encyclopedia of the Social \& Behavioral Sciences}, pages 7245--7248. Pergamon, Oxford.

\bibitem[{Lena(2021)}]{biosocial}
Palaniyappan Lena. 2021.
\newblock \href {https://www.proquest.com/scholarly-journals/more-than-biomarker-could-language-be-biosocial/docview/2567801968/se-2} {More than a biomarker: could language be a biosocial marker of psychosis?}
\newblock \emph{NPJ Schizophrenia}, 7(1).
\newblock Copyright - © The Author(s) 2021. This work is published under http://creativecommons.org/licenses/by/4.0/ (the “License”). Notwithstanding the ProQuest Terms and Conditions, you may use this content in accordance with the terms of the License; Last updated - 2023-02-22.

\bibitem[{Losada et~al.(2017)Losada, Crestani, and Parapar}]{eRisk2017}
David~E Losada, Fabio Crestani, and Javier Parapar. 2017.
\newblock erisk 2017: Clef lab on early risk prediction on the internet: experimental foundations.
\newblock In \emph{Experimental IR Meets Multilinguality, Multimodality, and Interaction: 8th International Conference of the CLEF Association, CLEF 2017, Dublin, Ireland, September 11--14, 2017, Proceedings 8}, pages 346--360. Springer.

\bibitem[{Losada et~al.(2018)Losada, Crestani, and Parapar}]{eRisk2018}
David~E Losada, Fabio Crestani, and Javier Parapar. 2018.
\newblock Overview of erisk: early risk prediction on the internet.
\newblock In \emph{Experimental IR Meets Multilinguality, Multimodality, and Interaction: 9th International Conference of the CLEF Association, CLEF 2018, Avignon, France, September 10-14, 2018, Proceedings 9}, pages 343--361. Springer.

\bibitem[{Mohammad(2011)}]{mohammad-2011-upon}
Saif Mohammad. 2011.
\newblock \href {https://aclanthology.org/W11-1514} {From once upon a time to happily ever after: Tracking emotions in novels and fairy tales}.
\newblock In \emph{Proceedings of the 5th {ACL}-{HLT} Workshop on Language Technology for Cultural Heritage, Social Sciences, and Humanities}, pages 105--114, Portland, OR, USA. Association for Computational Linguistics.

\bibitem[{Mohammad(2023)}]{mohammad2020practicaleacl}
Saif Mohammad. 2023.
\newblock \href {https://aclanthology.org/2023.findings-eacl.136} {Best practices in the creation and use of emotion lexicons}.
\newblock In \emph{Findings of the Association for Computational Linguistics: EACL 2023}, pages 1825--1836, Dubrovnik, Croatia. Association for Computational Linguistics.

\bibitem[{Mohammad(2018)}]{vad-acl2018}
Saif~M. Mohammad. 2018.
\newblock Obtaining reliable human ratings of valence, arousal, and dominance for 20,000 english words.
\newblock In \emph{Proceedings of The Annual Conference of the Association for Computational Linguistics (ACL)}, Melbourne, Australia.

\bibitem[{Mohammad(2022)}]{Mohammad22AER}
Saif~M. Mohammad. 2022.
\newblock Ethics sheet for automatic emotion recognition and sentiment analysis.
\newblock \emph{To Appear in Computational Linguistics}.

\bibitem[{Osatuke et~al.(2007)Osatuke, Mosher, Goldsmith, Stiles, Shapiro, Hardy, and Barkham}]{Osatuke}
Katerine Osatuke, James~K. Mosher, Jacob~Z. Goldsmith, William~B. Stiles, David~A. Shapiro, Gillian~E. Hardy, and Michael Barkham. 2007.
\newblock \href {https://doi.org/https://doi.org/10.1002/jclp.20338} {Submissive voices dominate in depression: Assimilation analysis of a helpful session}.
\newblock \emph{Journal of Clinical Psychology}, 63(2):153--164.

\bibitem[{Phillips et~al.(2002)Phillips, Phillips, Bull, Adams, and Fraser}]{Phillips}
Louise~H Phillips, Louise~H Phillips, Rebecca Bull, Ewan Adams, and Lisa Fraser. 2002.
\newblock \href {https://doi.org/10.1037/1528-3542.2.1.12} {Positive mood and executive function: evidence from stroop and fluency tasks}.
\newblock \emph{Emotion (Washington, D.C.)}, 2(1):12—22.

\bibitem[{Pollack(2005)}]{pollack2005comorbid}
Mark~H Pollack. 2005.
\newblock Comorbid anxiety and depression.
\newblock \emph{Journal of Clinical Psychiatry}, 66:22.

\bibitem[{Pugach et~al.(2023)Pugach, May, and Wisco}]{Pugach2023}
Cameron~P. Pugach, Casey~L. May, and Blair~E. Wisco. 2023.
\newblock \href {https://doi.org/10.1002/jts.22928} {Positive emotion in posttraumatic stress disorder: A global or context‐specific problem?}
\newblock \emph{Journal of Traumatic Stress}, 36(2):444--456.

\bibitem[{Ranney et~al.(2020)Ranney, Behar, and Yamasaki}]{RANNEY2020101542}
Rachel~M. Ranney, Evelyn Behar, and Alissa~S. Yamasaki. 2020.
\newblock \href {https://doi.org/https://doi.org/10.1016/j.jbtep.2019.101542} {Affect variability and emotional reactivity in generalized anxiety disorder}.
\newblock \emph{Journal of Behavior Therapy and Experimental Psychiatry}, 68:101542.

\bibitem[{Reagan et~al.(2016)Reagan, Mitchell, Kiley, Danforth, and Dodds}]{emotionarcs}
Andrew~J. Reagan, Lewis Mitchell, Dilan Kiley, Christopher~M. Danforth, and Peter~S. Dodds. 2016.
\newblock \href {https://www.proquest.com/scholarly-journals/emotional-arcs-stories-are-dominated-six-basic/docview/1865288690/se-2} {The emotional arcs of stories are dominated by six basic shapes}.
\newblock \emph{EPJ Data Science}, 5(1):1--12.
\newblock Copyright - EPJ Data Science is a copyright of Springer, 2016; Last updated - 2017-02-06.

\bibitem[{Russell(2003)}]{russell2003core}
James~A Russell. 2003.
\newblock Core affect and the psychological construction of emotion.
\newblock \emph{Psychological review}, 110(1):145.

\bibitem[{Seabrook et~al.(2018)Seabrook, Kern, Fulcher, and Rickard}]{seabrook}
Elizabeth~M Seabrook, Margaret~L Kern, Ben~D Fulcher, and Nikki~S Rickard. 2018.
\newblock \href {https://doi.org/10.2196/jmir.9267} {Predicting depression from language-based emotion dynamics: Longitudinal analysis of facebook and twitter status updates}.
\newblock \emph{J Med Internet Res}, 20(5):e168.

\bibitem[{Silk et~al.(2011)Silk, Forbes, Whalen, Jakubcak, Thompson, Ryan, Axelson, Birmaher, and Dahl}]{depressedYouth}
Jennifer~S. Silk, Erika~E. Forbes, Diana~J. Whalen, Jennifer~L. Jakubcak, Wesley~K. Thompson, Neal~D. Ryan, David~A. Axelson, Boris Birmaher, and Ronald~E. Dahl. 2011.
\newblock \href {https://doi.org/https://doi.org/10.1016/j.jecp.2010.10.007} {Daily emotional dynamics in depressed youth: A cell phone ecological momentary assessment study}.
\newblock \emph{Journal of Experimental Child Psychology}, 110(2):241--257.
\newblock Special Issue: Assessment of Emotion in Children and Adolescents.

\bibitem[{Sosa-Hernandez et~al.(2022)Sosa-Hernandez, Wilson, and Henderson}]{Sosa}
Linda Sosa-Hernandez, McLennon Wilson, and Heather~A Henderson. 2022.
\newblock \href {https://doi.org/10.1037/emo0001155} {Emotion dynamics among preadolescents getting to know each other: Dyadic associations with shyness}.
\newblock \emph{Emotion (Washington, D.C.)}.

\bibitem[{Sperry et~al.(2020)Sperry, Walsh, and Kwapil}]{sperry}
Sarah~H. Sperry, Molly~A. Walsh, and Thomas~R. Kwapil. 2020.
\newblock \href {https://doi.org/https://doi.org/10.1016/j.jad.2019.09.076} {Emotion dynamics concurrently and prospectively predict mood psychopathology}.
\newblock \emph{Journal of Affective Disorders}, 261:67--75.

\bibitem[{Stasak et~al.(2016)Stasak, Epps, Cummins, and Goecke}]{StasakEppsCumminsetal2016}
Brian Stasak, Julien Epps, Nicholas Cummins, and Roland Goecke. 2016.
\newblock \href {https://doi.org/10.21437/interspeech.2016-867} {An investigation of emotional speech in depression classification}.
\newblock In \emph{Understanding speech processing in humans and machines: 17th Annual Conference of the International Speech Communication Association (INTERSPEECH 2016), San Francisco, California, USA, 8-12 September 2016; Volume 1}.

\bibitem[{Stickley et~al.(2018)Stickley, Koyanagi, Takahashi, Ruchkin, Inoue, Yazawa, and Kamio}]{STICKLEY2018317}
Andrew Stickley, Ai~Koyanagi, Hidetoshi Takahashi, Vladislav Ruchkin, Yosuke Inoue, Aki Yazawa, and Yoko Kamio. 2018.
\newblock \href {https://doi.org/https://doi.org/10.1016/j.psychres.2018.05.004} {Attention-deficit/hyperactivity disorder symptoms and happiness among adults in the general population}.
\newblock \emph{Psychiatry Research}, 265:317--323.

\bibitem[{Suhavi et~al.(2022)Suhavi, Singh, Arora, Shrivastava, Singh, Shah, and Kumaraguru}]{_Singh_Arora_Shrivastava_Singh_Shah_Kumaraguru_2022}
Suhavi, Asmit~Kumar Singh, Udit Arora, Somyadeep Shrivastava, Aryaveer Singh, Rajiv~Ratn Shah, and Ponnurangam Kumaraguru. 2022.
\newblock \href {https://doi.org/10.1609/icwsm.v16i1.19368} {Twitter-stmhd: An extensive user-level database of multiple mental health disorders}.
\newblock \emph{Proceedings of the International AAAI Conference on Web and Social Media}, 16(1):1182--1191.

\bibitem[{Teodorescu et~al.(2023)Teodorescu, Fyshe, and Mohammad}]{teodorescu2023utterance}
Daniela Teodorescu, Alona Fyshe, and Saif Mohammad. 2023.
\newblock \href {https://doi.org/10.18653/v1/2023.wassa-1.35} {Utterance emotion dynamics in children{'}s poems: Emotional changes across age}.
\newblock In \emph{Proceedings of the 13th Workshop on Computational Approaches to Subjectivity, Sentiment, {\&} Social Media Analysis}, pages 401--415, Toronto, Canada. Association for Computational Linguistics.

\bibitem[{Teodorescu and Mohammad(2023)}]{teodorescu2023emoarc}
Daniela Teodorescu and Saif Mohammad. 2023.
\newblock Evaluating emotion arcs across languages: Bridging the global divide in sentiment analysis.
\newblock In \emph{The 2023 Conference on Empirical Methods in Natural Language Processing}.

\bibitem[{Vishnubhotla and Mohammad(2022{\natexlab{a}})}]{vishnubhotla-mohammad-2022-tusc}
Krishnapriya Vishnubhotla and Saif~M. Mohammad. 2022{\natexlab{a}}.
\newblock \href {https://aclanthology.org/2022.lrec-1.442} {{Tweet Emotion Dynamics}: Emotion word usage in tweets from {US} and {C}anada}.
\newblock In \emph{Proceedings of the Thirteenth Language Resources and Evaluation Conference}, pages 4162--4176, Marseille, France. European Language Resources Association.

\bibitem[{Vishnubhotla and Mohammad(2022{\natexlab{b}})}]{VM2022-TED}
Krishnapriya Vishnubhotla and Saif~M. Mohammad. 2022{\natexlab{b}}.
\newblock Tweet emotion dynamics: Emotion word usage in tweets from us and canada.
\newblock In \emph{Proceedings of the Thirteenth International Conference on Language Resources and Evaluation (LREC 2022)}, Marseille, France.

\end{thebibliography}
\bibliographystyle{acl_natbib}

\appendix

\noindent \textbf{APPENDIX}
\section{Twitter-STMHD Dataset}
\label{app:twitter_dataset}
\citet{_Singh_Arora_Shrivastava_Singh_Shah_Kumaraguru_2022} 
created a regular expression pattern 
\DT{to identify posts which contained}
%containing 
a self-disclosure of a diagnosis and the diagnosis name (using a lexicon of common synonyms, abbreviations, etc.) such as `diagnosed with X'. 
They collected a large set of tweets using the regex. 
This resulted in a preliminary dataset of users with potential MHC diagnoses.
% however, a large number of false positives may have occurred as a result of matching the regular expression 
To handle false positives (e.g., `my family member has been diagnosed with X', or `I was not diagnosed with X'),
\DT{the dataset was split into two non-overlapping parts, one of which was manually annotated, and the other using an updated and high-precision regex.}
\DT{In the part that was annotated by hand, each tweet was annotated by two members of the team. A user was only included in the dataset if both annotations were positive as self-disclosing for a particular class. A licensed clinical psychologist % helped verify a sample of 500 tweets from this part of the dataset. Comparing the authors' annotations to those of the psychologist's, the authors annotated 
found the 500-tweet sample to be 99.2\% accurate.
The manual annotations were used to refine the regular expressions and diagnosis name lexicon. This updated search pattern was applied to the other dataset split.
To verify the quality of the updated regex, the authors applied it to the manually annotated dataset split. When considering the manual annotations as correct, the regex was found to be  94\% accurate.}

\section{Statistical Results}

\subsection{Levene's Test}
\label{appendix:levenes}

Levene’s test indicated that the assumption for homogeneity of variance was violated for the effect of diagnosis on all UED metrics (emotional average, emotional variability, rise rate, and recovery rate) across all three \DT{dimensional} emotions (valence, arousal, and dominance).
We show the results in Table \ref{tab:levens}.

% emotional average 
% (\textit{F}(7, 15027)=60.84, \textit{p}<.001), emotional variability 
% (\textit{F}(7, 15027)=71.53, \textit{p}<.001),
% rise-rate (\textit{F}(7, 15027)=82.73, \textit{p}<.001), 
% and recovery rate 
% (\textit{F}(7, 15027)=88.24, \textit{p}<.001).
% (\textit{F}(7, 14654)=59.913, \textit{p}<.001),
% emotional variability (\textit{F}(7, 14654)=67.98, \textit{p}<.001),
% rise-rate (\textit{F}(7, 14652)=76.85, \textit{p}<.001), 
% and recovery rate (\textit{F}(7, 14652)=82.41, \textit{p}<.001).

\begin{table*}[!ht]
\centering
{\small
\begin{tabular}
{llrrrr}
\hline
\multicolumn{1}{l}{\textbf{Emotion}} &
\multicolumn{1}{l}{\textbf{UED Metric}} &
\multicolumn{1}{l}{\textbf{df1}} &
\multicolumn{1}{l}{\textbf{df2}} &
\multicolumn{1}{l}{\textbf{F-statistic}} &
\multicolumn{1}{l}{\textbf{P-value}} \\ \hline
% Valence & emotional average & 7 & 14654 & 59.91 & \textit{p}<.001 \\
% & emotional variability & 7 & 14654 & 67.98 & \textit{p}<.001\\
% & rise rate & 7 & 14654 & 76.85 & \textit{p}<.001\\
% & recovery rate & 7 & 14652 & 82.41 & \textit{p}<.001 \\ 

% Arousal & emotional average & 7 & 14654 & 37.69 & \textit{p}<.001 \\
% & emotional variability & 7 & 14654 & 23.09 & \textit{p}<.001\\
% & rise rate & 7 & 14651 & 36.34 & \textit{p}<.001\\
% & recovery rate & 7 & 14650 & 43.66 & \textit{p}<.001 \\ 

% Dominance & emotional average & 7 & 14654 & 60.93 & \textit{p}<.001 \\
% & emotional variability & 7 & 14654 & 73.45 & \textit{p}<.001\\
% & rise rate & 7 & 14646 & 37.49 & \textit{p}<.001\\
% & recovery rate & 7 & 14650 & 40.52 & \textit{p}<.001 \\ 

Valence & average emotion & 7 & 14158 & 60.50 & \textit{p}$<$.001 \\
& emotional variability & 7 & 14158 & 66.33 & \textit{p}$<$.001\\
& rise rate & 7 & 14156 & 77.35 & \textit{p}$<$.001\\
& recovery rate & 7 & 14156 & 72.58 & \textit{p}$<$.001 \\ 

Arousal & average emotion & 7 & 14158 & 37.60 & \textit{p}$<$.001 \\
& emotional variability & 7 & 14158 & 22.38 & \textit{p}$<$.001\\
& rise rate & 7 & 14155 & 34.76 & \textit{p}$<$.001\\
& recovery rate & 7 & 14155 & 41.28 & \textit{p}$<$.001 \\  

Dominance & average emotion & 7 & 14158 & 61.86 & \textit{p}$<$.001 \\
& emotional variability & 7 & 14158 & 72.21 & \textit{p}$<$.001\\
& rise rate & 7 & 14150 & 37.69 & \textit{p}$<$.001\\
& recovery rate & 7 & 14154 & 35.64 & \textit{p}$<$.001 \\ 
\hline
\end{tabular}
}
% \vspace*{-1mm}
\caption{The degrees of freedom, F-statistic, and p-value in Levene's test of Homogeneity of Variances for each UED metric and emotion.}
\vspace*{3mm}
\label{tab:levens}
\end{table*}

\subsection{Welch's ANOVA}

In Table \ref{tab:welchs_valence_arousal_dominance} we show the results for Welch's ANOVA for valence, arousal, and dominance.

\begin{table*}[!ht]
\centering
{\small
\begin{tabular}
{llrrrrr}
\hline
\multicolumn{1}{l}{\textbf{Emotion}} &
\multicolumn{1}{l}{\textbf{UED Metric}} &
\multicolumn{1}{l}{\textbf{df1}} &
\multicolumn{1}{l}{\textbf{df2}} &
\multicolumn{1}{l}{\textbf{F-statistic}} &
\multicolumn{1}{l}{\textbf{P-value}} &
\multicolumn{1}{l}{\textbf{Effect Size (\textit{est $\omega$}\textsuperscript{2})}}\\ \hline
% Arousal & emotional average & 7 & 1103.36 & 32.10 & \textit{p}<.001 & 0.0146\\
% & emotional variability & 7 & 1107.28 & 67.41 & \textit{p}<.001 & 0.0307\\
% & rise rate & 7 & 1104.64 & 2.70 & \textit{p}<.001 & 0.0008\\
% & recovery rate & 7 & 1105.81 & 4.52 & \textit{p}<.001 & 0.0017\\ 

% Dominance & emotional average & 7 & 1097.65 & 55.51 & \textit{p}<.001 & 0.0254\\
% & emotional variability & 7 & 1102.04 & 41.97 & \textit{p}<.001 & 0.0192\\
% & rise rate & 7 & 1102.82 & 6.14 & \textit{p}<.001 & 0.0024\\
% & recovery rate & 7 & 1102.27 & 7.56 & \textit{p}<.001 & 0.0031\\ 

Valence & average emotion & 7 & 1021.65 & 14.79 & \textit{p}$<$.001 & 0.0068 \\
& emotional variability & 7 & 1021.20 & 70.30 & \textit{p}$<$.001 & 0.0331 \\
& rise rate & 7 & 1026.32 & 9.93 & \textit{p}$<$.001 & 0.0044 \\
& recovery rate & 7 & 1023.62 & 8.86 & \textit{p}$<$.001 & 0.0039 \\

Arousal & average emotion & 7 & 1024.41 & 33.24 & \textit{p}$<$.001 & 0.0157\\
& emotional variability & 7 & 1029.77 & 66.23 & \textit{p}$<$.001 & 0.0312\\
& rise rate & 7 & 1025.85 & 2.84 & \textit{p}$=$.006 & 0.0009\\
& recovery rate & 7 & 1026.95 & 5.19 & \textit{p}$<$.001 & 0.0021\\

Dominance & average emotion & 7 & 1020.10 & 56.69 & \textit{p}$<$.001 & 0.0268\\
& emotional variability & 7 & 1023.12 & 40.50 & \textit{p}$<$.001 & 0.0191\\
& rise rate & 7 & 1025.35 & 6.31 & \textit{p}$<$.001 & 0.0026\\
& recovery rate & 7 & 1022.99 & 9.94 & \textit{p}$<$.001 & 0.0044\\ 
\hline
\end{tabular}
}
% \vspace*{-1mm}
\caption{The degrees of freedom (for the numerator and denominator), F-statistic, p-value, and effect size in Welch's ANOVA test for differences between groups, for each UED metric and emotion (valence, arousal, and dominance).% ``emo.'' is an abbreviation for emotional.
}
\vspace*{3mm}
\label{tab:welchs_valence_arousal_dominance}
\end{table*}

\section{Data Descriptives for Distributions of Social Aspects}

% \textbf{Number of People Following}: The distribution for the number of following had the following characteristics: average = 775.46, standard deviation =  884.13, median = 475.00, 25\% percentile = 236.00, and 75\% percentile = 935.00.

% The distribution for the average number of likes a user received had the following characteristics: average = 2.05, standard deviation = 8.21, median = 1.20, 25\% percentile = 0.68, and 75\% percentile = 2.11.
In Table \ref{tab:social_descriptive} we show some key details about the distribution of 
the %three 
\textit{popularity} aspect on Twitter: 
%number of people following,
%number of followers, 
average likes per Tweeter.

\begin{table*}[!ht]
\centering
{\small
\begin{tabular}
{lrrrrr}
\hline
\multicolumn{1}{l}{\textbf{Popularity Aspect}} &
\multicolumn{1}{l}{\textbf{Average}} &
\multicolumn{1}{l}{\textbf{Std. Dev.}} &
\multicolumn{1}{l}{\textbf{25th Percentile}} & 
\multicolumn{1}{l}{\textbf{Median}} &
\multicolumn{1}{l}{\textbf{75th Percentile}}\\ \hline
% No. Following & 775.46 &  884.13 & 236.00 & 475.00 & 935.00 \\
% No. Followers & 632.75 & 825.56 & 139.00 & 321.00 & 756.00 \\
% Avg. Likes & 2.06 & 8.21 & 0.68 & 1.20 & 2.11 \\

% No. Following & 771.19 & 881.22 & 235.5 & 472.0 & 928.00 \\
% No. Followers & 626.89 &  821.42 & 138.00 & 317.00 & 748.00 \\
Avg.\@ Likes per Tweeter & 2.06 & 8.33 & 0.67 & 1.20 & 2.11 \\
\hline
\end{tabular}
}
% \vspace*{-1mm}
\caption{Key statistics of the distribution of the \textit{popularity} aspect: Avg.\@ Likes per Tweeter.} 
% number of people following, 
% number of followers,
% \DT{and}
% average likes.}
% \vspace*{-3mm}
\label{tab:social_descriptive}
\end{table*}

\section{UED Across Average Likes per Tweeter: Valence}
\label{app:ued_likes_valence}
\begin{figure*}[h]
    \centering
    % \begin{subfigure}[b]{0.45\textwidth}
    % \centering
    % \subcaption{}
    % \includegraphics[width=\textwidth]{images2/Average_Valence_avg_likes_new_chart_Oct22.png}
    % % \caption{}
    % \label{fig:emo_mean_val}
    % \end{subfigure}
    % \hfill
    % \begin{subfigure}[b]{0.45\textwidth}
    % \centering
    % \subcaption{}
    % \includegraphics[width=\textwidth]{images2/Varability_Valence_avg_likes_new_chart_Oct22.png}
    % % \caption{}
    % \label{fig:emo_std_val}
    % \end{subfigure}
    % %\caption{Recovery rate in poems across grades. 
    % %The horizontal dashed lines represent values in poems written by adults.
    % %, and per gender across grades.
    % %}
    % \vskip\baselineskip
    % \vspace*{-8mm}
    % \centering
    \begin{subfigure}[b]{0.45\textwidth}
    \centering
    \subcaption{}
    \includegraphics[width=\textwidth]{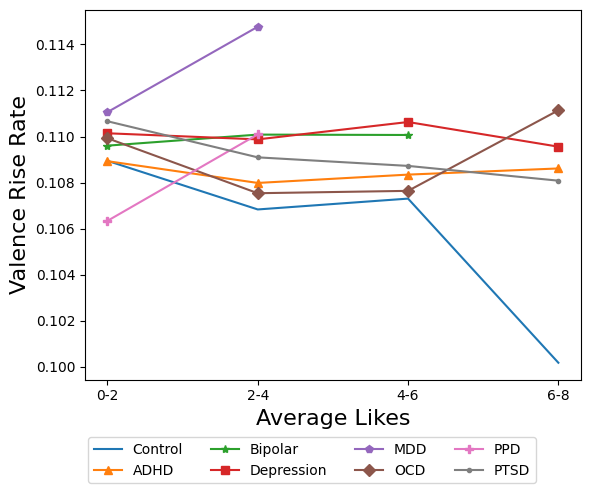}
    % \caption{}
    \label{fig:rise_rate_val}
    \end{subfigure}
    \hfill
    \begin{subfigure}[b]{0.45\textwidth}
    \centering
    \subcaption{}
    \includegraphics[width=\textwidth]{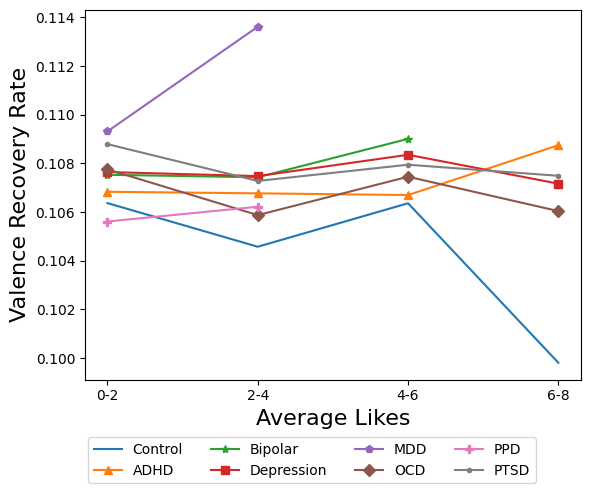}
    % \caption{}
    \label{fig:recovery_rate_val}
    \end{subfigure}
    \vspace*{-3mm}
    \caption{%UED metrics for 
    \textbf{Valence} rise and recovery rates across % various 
    levels of user \textit{popularity} on Twitter (average \#likes on posts). % received per post.
    }
    \label{fig:ued_val_avg_likes2}
\end{figure*}

\DTtwo{In Figure \ref{fig:ued_val_avg_likes2}, we show two UED metrics (e.g., rise rate and recovery rate) for valence, across various bins of average likes received for each group.}

\section{UED Across Average Likes per Tweeter: Arousal}
\label{app:ued_likes_arousal}

\begin{figure*}[h]
    \centering
    \begin{subfigure}[b]{0.45\textwidth}
    \centering
    \subcaption{}
    \includegraphics[width=\textwidth]{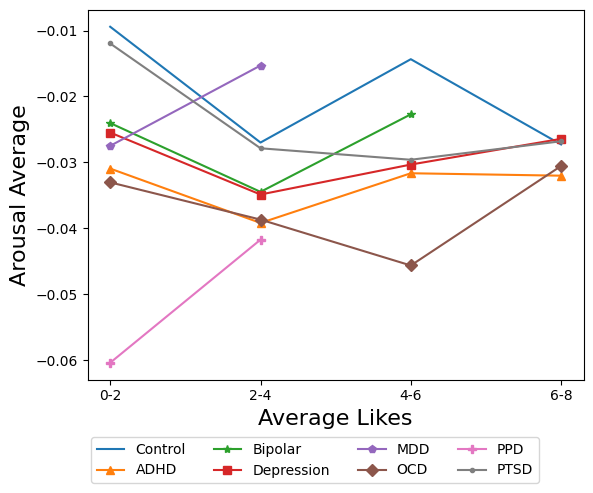}
    % \captionsetup[subfigure]{position=top, labelfont=bf,textfont=normalfont,singlelinecheck=off,justification=raggedright}
    \label{fig:emo_mean_arous}
    \end{subfigure}
    \hfill
    \begin{subfigure}[b]{0.45\textwidth}
    \centering
    \subcaption{}
    \includegraphics[width=\textwidth]{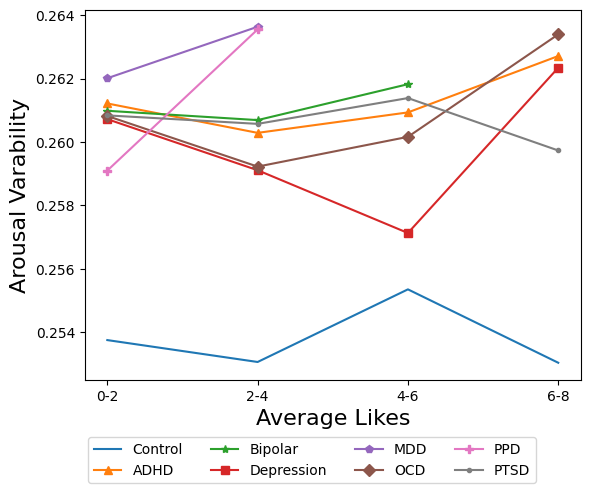}
    %\caption{Anger, fear, joy and sadness}
    \label{fig:emo_std_arous}
    \end{subfigure}
    %\caption{Recovery rate in poems across grades. 
    %The horizontal dashed lines represent values in poems written by adults.
    %, and per gender across grades.
    %}
    \vskip\baselineskip
    \vspace*{-8mm}
    \centering
    \begin{subfigure}[b]{0.45\textwidth}
    \centering
    \subcaption{}
\includegraphics[width=\textwidth]{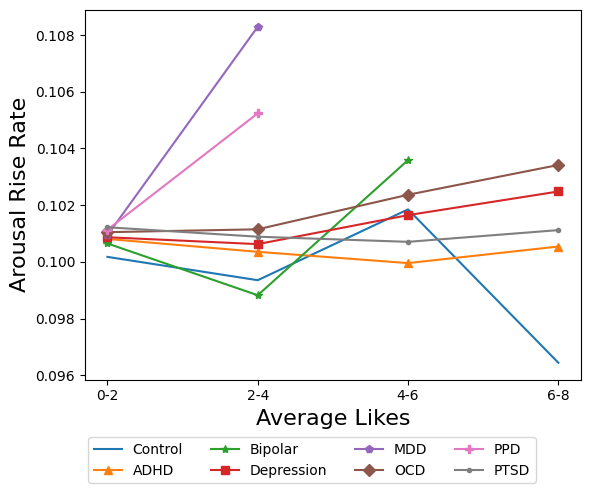}
    \label{fig:rise_rate_arous}
    \end{subfigure}
    \hfill
    \begin{subfigure}[b]{0.45\textwidth}
    \centering
    \subcaption{}
    \includegraphics[width=\textwidth]{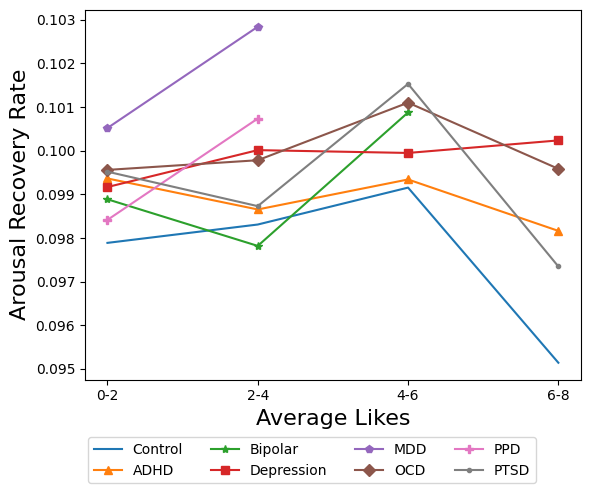}
    %\caption{Anger, fear, joy and sadness}
    \label{fig:recovery_rate_arous}
    \end{subfigure}
    \vspace*{-5mm}
    \caption{UED metrics for \textbf{arousal} across various levels of user \textit{popularity} on Twitter (average \#likes on posts). % received per post.
    }
    \label{fig:ued_arous_avg_likes}
\end{figure*}

\DT{In Figure \ref{fig:ued_arous_avg_likes}, we show the four UED metrics (e.g., average emotion intensity, emotional variability, rise rate, and recovery rate) for arousal, across various bins of average likes received for each group.}

\noindent \textit{Average Arousal:} In Section \ref{sec:results}, we showed that average arousal is markedly lower for ADHD, Bipolar, OCD, PPD, and Depression. From Figure \ref{fig:ued_arous_avg_likes} (a), we observe that the average arousal for these MHCs is markedly lower than control for various Average-Likes bins.
 (We also observe slightly smaller differences for Depression and OCD when average likes is in the [6,8) range, perhaps pointing to the potential for average arousal to be influenced when a post is considered quite \textit{popular}.)
\\[4pt] 
\noindent \textit{Arousal Variability:} In Section \ref{sec:results}, we showed that arousal variability is markedly higher for all MHCs. From Figure \ref{fig:ued_arous_avg_likes} (b), we observe that the arousal variability for these MHCs is markedly higher than control for various Average-Likes bins. (We also observe markedly lower variability for control when average likes is in the [6,8) range.)\\[4pt] 
\noindent \textit{Arousal Rise Rate:} In Section \ref{sec:results}, we showed that arousal rise rate is markedly higher for PTSD. From Figure \ref{fig:ued_arous_avg_likes} (c), we observe that the arousal rise rate for PTSD does cross the control line at bin 3 with [4-6) likes. This points to the potential for arousal rise rate to be slightly influenced by a popularity aspect such as the average number of likes received.
(We also observe markedly lower arousal rise rate for control when average user likes is in the [6,8) range.)\\[4pt] 
\noindent \textit{Arousal Recovery Rate:} In Section \ref{sec:results}, we showed that arousal recovery rate is markedly higher for ADHD, MDD, OCD, PTSD, and Depression. From Figure \ref{fig:ued_arous_avg_likes} (d), we observe that the arousal recovery rate for these MHCs is markedly higher than control for various Average-Likes bins (slightly less for ADHD). (We also observe markedly lower 
arousal
recovery rate for control when average user likes is in the [6,8) range.) % \\[pt] 

Thus, overall, we observe that for the arousal UEDs which showed significant differences across MHCs and control, these differences still do appear (although slightly less for arousal rise and recovery rates) when accounting for popularity measures such as the average number of likes a user's posts get. 
% In Figure \ref{fig:ued_arous_avg_likes} we show the UED metrics for arousal across various levels of the popularity measure: average number of likes received on posts. Overall, it seems that 
%  UED are markedly different for MHCs compared to the control group even after accounting for the average number of likes. The lines for MHCs do not cross the control line for average arousal or arousal variability, and if they do (e.g., MDD for average arousal), there was no significant difference found in Section \ref{sec:results}.
%  For recovery rate, although bipolar does cross the control line, these diagnoses were not significantly different for arousal recovery rate (Section \ref{sec:results}).
%  However, for arousal rise rate the line for PTSD does cross the control line at bin 3 (e.g., 4--6 likes) and these diagnoses were significantly different in Section \ref{sec:results}. This points to the potential for arousal rise rate to be influenced by popularity measures.
%  Largely overall, there are still marked differences even after accounting for popularity.      
% the average number of likes received may influence UED metrics such as rise and recovery rate for arousal, where the lines for MHCs cross the control line.
% Otherwise, arousal variability and average arousal (somewhat less) still seem to be markedly different for the MHCs compared to the control group.

% State how the marked difference indicates that the difference exists even when one accounts for popularity

\section{UED Across Average Likes per Tweeter: Dominance}
\label{app:ued_likes_dominance}

\begin{figure*}[h]
    \centering
    \begin{subfigure}[b]{0.45\textwidth}
    \centering
    \subcaption{}
    \includegraphics[width=\textwidth]{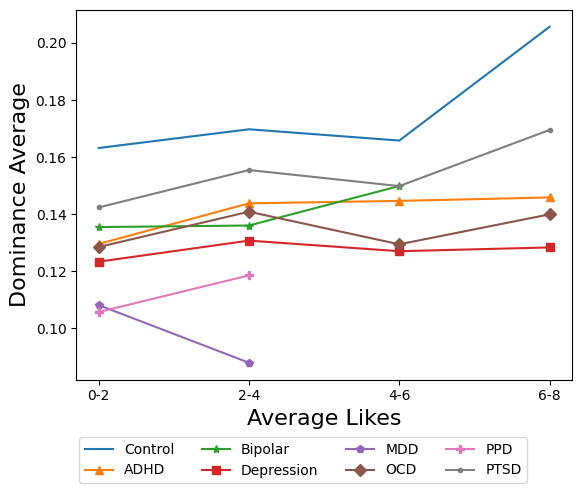}
    %\caption{Valence}
    \label{fig:emo_mean_dom}
    \end{subfigure}
    \hfill
    \begin{subfigure}[b]{0.45\textwidth}
    \centering
    \subcaption{}
    \includegraphics[width=\textwidth]{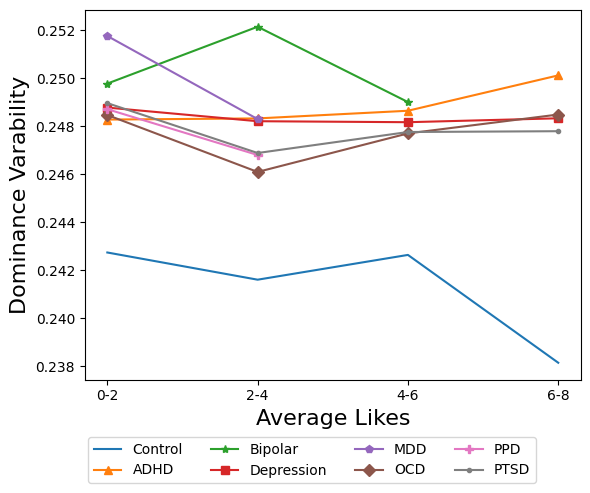}
    %\caption{Anger, fear, joy and sadness}
    \label{fig:emo_std_dom}
    \end{subfigure}
    %\caption{Recovery rate in poems across grades. 
    %The horizontal dashed lines represent values in poems written by adults.
    %, and per gender across grades.
    %}
    \vskip\baselineskip
    \vspace*{-8mm}
    \centering
    \begin{subfigure}[b]{0.45\textwidth}
    \centering
    \subcaption{}
    \includegraphics[width=\textwidth]{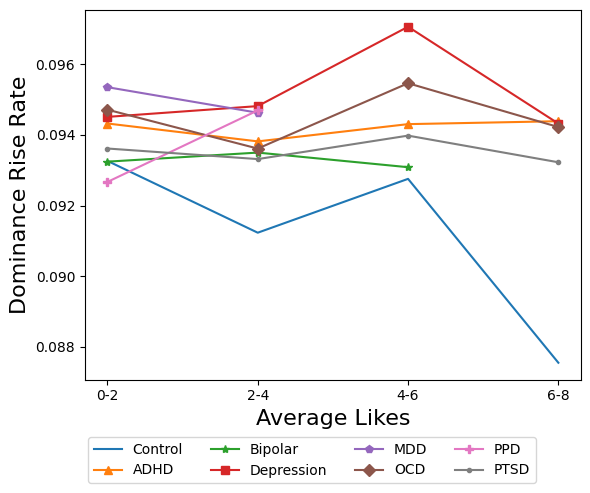}
    %\caption{Valence}
    \label{fig:rise_rate_dom}
    \end{subfigure}
    \hfill
    \begin{subfigure}[b]{0.45\textwidth}
    \centering
    \subcaption{}
    \includegraphics[width=\textwidth]{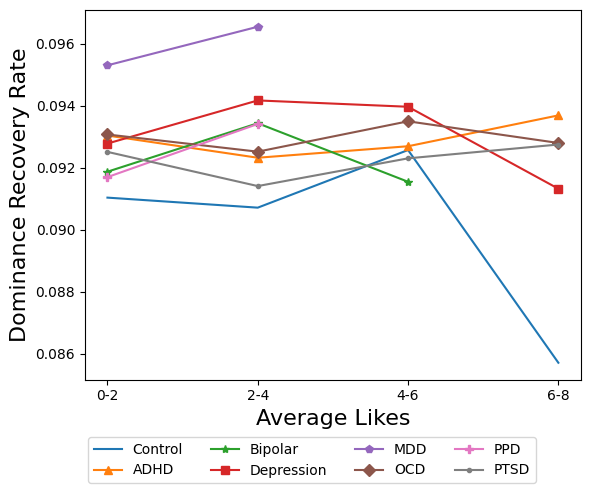}
    %\caption{Anger, fear, joy and sadness}
    \label{fig:recovery_rate_dom}
    \end{subfigure}
    \vspace*{-5mm}
    \caption{UED metrics for \textbf{dominance} across various levels of user \textit{popularity} on Twitter (average \#likes on posts). % received per post.
    }
    \label{fig:ued_dom_avg_likes}
\end{figure*}

\DT{In Figure \ref{fig:ued_dom_avg_likes}, we show the four UED metrics (e.g., average emotion intensity, emotional variability, rise rate, and recovery rate) for dominance, across various bins of average likes received for each group.}

\noindent \textit{Average Dominance:} In Section \ref{sec:results}, we showed that average dominance is markedly lower for all MHCs. From Figure \ref{fig:ued_dom_avg_likes} (a), we observe that the average dominance for these MHCs is markedly lower than control for various Average-Likes bins.\\[4pt] 
\noindent \textit{Dominance Variability:} In Section \ref{sec:results}, we showed that dominance variability is markedly higher for all MHCs. From Figure \ref{fig:ued_dom_avg_likes} (b), we observe that the dominance variability for these MHCs is markedly higher than control for various Average-Likes bins. (We also observe markedly lower variability for control when average likes is in the [6,8) range.)\\[4pt] 
\noindent \textit{Dominance Rise Rate:} In Section \ref{sec:results}, we showed that dominance rise rate is markedly higher for ADHD, OCD and Depression. From Figure \ref{fig:ued_dom_avg_likes} (c), we observe that the dominance rise rate for these MHCs is markedly higher than control for various Average-Likes bins. 
(We also observe markedly lower dominance rise rate for control when average user likes is in the [6,8) range.)\\[4pt] 
\noindent \textit{Dominance Recovery Rate:} In Section \ref{sec:results}, we showed that dominance recovery rate is markedly higher for ADHD, MDD, OCD, PTSD, and Depression. From Figure \ref{fig:ued_dom_avg_likes} (d), we observe that the dominance recovery rate for these MHCs is higher than control for various Average-Likes bins although not for ADHD, PTSD which intersect the control line at bin 3 with [4-6) likes. We also notice that the difference dominance recovery rate for OCD and control is less at bin 3 with [4-6) likes. This points to the potential that dominance recovery rate may be influenced by a popularity measure such as the average number of likes received.  
(We also observe markedly lower 
dominance recovery rate for control when average user likes is in the [6,8) range.)  \\[3pt] 
Thus, overall, we observe that for the dominance UEDs which showed significant differences across MHCs and control, these differences still do largely appear (although less for dominance recovery rates) when accounting for popularity measures such as the average number of likes a user's posts get. 
% \DT{In Figure \ref{fig:ued_val_avg_likes}, we show the four UED metrics (e.g., average emotion intensity, emotional variability, rise rate, and recovery rate) for valence, across various bins of average likes received for each group.}

% In Figure \ref{fig:ued_dom_avg_likes} we show the UED metrics for dominance across various levels of the popularity measure: average number of likes received on posts.
% The average emotion intensity for dominance and variability still appear to be markedly different for the MHCs compared to the control group even after accounting for average number of likes. We see no lines for MHCs crossing the control group. Dominance rise rate for PPD crosses the control line, however this pair was not significantly different in Section \ref{sec:results}.
% However, for dominance recovery rate the line for PTSD does cross the control line at bin 3 (e.g., 4--6 likes) and these diagnoses were significantly different in Section \ref{sec:results}. This points to the potential for dominance recovery rate to be influenced by popularity measures.
% Largely overall, there are still marked differences even after accounting for popularity.   
% Overall, it seems that the average number of likes received may influence UED metrics such as rise and recovery rate for dominance (somewhat less than for arousal), since the lines for MHCs cross the control line (espescially for recovery rate). 

% State how the marked difference indicates that the difference exists even when one accounts for popularity

\section{UED Mean Differences}
\label{app:mean_diff}
Table \ref{tab:mean_diff_avg_var} and Table \ref{tab:mean_diff_rise_recover} show the mean difference in UEDs between the control group and an MHC if the difference was statistically significant in the pairwise comparison shown in Section \ref{sec:results}.

\begin{table*}[t]
\centering
{\small
\begin{tabular}{llrrrrrr}
%{0.9\textwidth}{|*{13}{p{1cm}|}}
\hline
%\multicolumn{1}{l}{\textbf{\backslashbox{LowerText}{UpperText}}} &
&  & %\multicolumn{1}{l|}{\textbf{\diagbox[width=\dimexpr \textwidth/4+2\tabcolsep\relax,, height=0.7cm]{\\MHC--}{UED}}} &
% \multicolumn{3}{p{2.5cm}}{\centering \textbf{Average Emotion}} &
% \multicolumn{3}{p{2.5cm}}{\textbf{Emotion Variability}} &
% \multicolumn{3}{p{2.5cm}}{\textbf{Rise Rate}} &
% \multicolumn{3}{p{2.5cm}}{\textbf{Recovery Rate}}
  \multicolumn{3}{c}{\textbf{Average Emotion}} & 
  \multicolumn{3}{c}{\textbf{Emotion Variability}} \\
  % \multicolumn{3}{l}{\textbf{Rise Rate}} & 
  % \multicolumn{3}{l}{\textbf{Recovery Rate}} \\ 
 % & & %\multicolumn{1}{l|}{\textbf{Control}} & 
  %\multicolumn{3}{l}{\textbf{Emotion}} & 
  %\multicolumn{3}{l}{\textbf{Variability}} & 
  % \multicolumn{3}{l}{\textbf{}}& 
  % \multicolumn{3}{l}{\textbf{Rate}}\\
 \hline
\textbf{Dataset} &\textbf{MHC--Control} & % &\multicolumn{1}{l|}{Emotion Dimension} & 
 \multicolumn{1}{c}{V} & \multicolumn{1}{c}{A} & \multicolumn{1}{c}{D} & 
 \multicolumn{1}{c}{V} & \multicolumn{1}{c}{A} & \multicolumn{1}{c}{D} % &
 %\multicolumn{1}{l}{V} & \multicolumn{1}{l}{A} & \multicolumn{1}{l}{D} &
 %\multicolumn{1}{l}{V} & \multicolumn{1}{l}{A} & \multicolumn{1}{l}{D}
 \\ \hline
 % & \multicolumn{1}{l|}{} & \multicolumn{1}{l|}{} &  & \multicolumn{1}{l|}{} & \multicolumn{1}{l|}{} &  & \multicolumn{1}{l|}{} & \multicolumn{1}{l|}{} &  & \multicolumn{1}{l|}{} & \multicolumn{1}{l|}{} &  \\
%& \multicolumn{1}{p{0.8cm}}{\textbf{V}} & & & & & \\
Twitter-STMHD & ADHD--control & -0.018 & -0.021 & -0.030 & 0.008 & 0.007 & 0.006 \\ % &  -- & -- & -0.001 & --  & -0.001 & $\uparrow$\\ % (-.011, -.006)
& Bipolar--control & -- & -0.013 & -0.029 & 0.015 & 0.007 & 0.008 \\ %& -- & -- & --   & -0.002 & -- & --\\ %(-.018,-.011)

& MDD--control & -0.027 & -- & -0.059 & 0.017 & 0.009 & 0.009 \\ % & -0.004 & -- & -- & -0.004 & -0.003 & $\uparrow$\\ %(-.023,-.012)
& OCD--control & -- & -0.023 & -0.033 & 0.010 & 0.007 & 0.005 \\ % & -- & -- & -0.002 &  -- & -0.002 & $\uparrow$\\ %(-.0136,-.007)
& PPD--control & -- & -0.046 & -0.051 &  -- & 0.006 & 0.006  \\ %& -- & -- & -- &  -- & -- & --\\ %(-.013,.002)
& PTSD--control & -0.023 & -- & -0.018 & 0.015 & 0.007 & 0.006 \\ %& -0.002 & -0.001 & -- & -0.002 & -0.001 & $\uparrow$ \\ %(-.018, -.013)

& Depression--control & -- & -0.015 & -0.040 & 0.013 & 0.007 & 0.006 \\ %& -0.002 & -- & -0.002 & -0.002 & -0.001 & $\uparrow$ \\ %(-.015,-.011)
Reddit eRisk & \DT{Depression--control} & -- & -- & -0.041 & 0.014 & -- & 0.006 \\ %& -- & -- & -- &  -- & -- & -- \\

    \hline
%\end{tabulary} 
\end{tabular}
}
% \vspace*{-1mm}
\caption{ % \textbf{Valence (V), Arousal (A), Dominance (D)}: 
The mean difference in UED metrics % across MHC groups compared to 
between each MHC group and the control if there was a significant difference between groups. 
%is indicated by an arrow; arrow direction indicates the direction of the difference. %; otherwise the cell has a dash. 
E.g., $-0.018$ for ADHD--control and average emotion `V' %column 
means that the ADHD group has significantly lower average valence by $0.018$ than the control group.
}
% \vspace*{-3mm}
\label{tab:mean_diff_avg_var}
\end{table*}

\begin{table*}[t]
\centering
{\small
\begin{tabular}{llrrrrrr}
%{0.9\textwidth}{|*{13}{p{1cm}|}}
\hline
%\multicolumn{1}{l}{\textbf{\backslashbox{LowerText}{UpperText}}} &
&  & %\multicolumn{1}{l|}{\textbf{\diagbox[width=\dimexpr \textwidth/4+2\tabcolsep\relax,, height=0.7cm]{\\MHC--}{UED}}} &
% \multicolumn{3}{p{2.5cm}}{\centering \textbf{Average Emotion}} &
% \multicolumn{3}{p{2.5cm}}{\textbf{Emotion Variability}} &
% \multicolumn{3}{p{2.5cm}}{\textbf{Rise Rate}} &
% \multicolumn{3}{p{2.5cm}}{\textbf{Recovery Rate}}
  % \multicolumn{3}{l}{\textbf{Average Emotion}} & 
  % \multicolumn{3}{l}{\textbf{Emotion Variability}} & 
  \multicolumn{3}{c}{\textbf{Rise Rate}} & 
  \multicolumn{3}{c}{\textbf{Recovery Rate}} \\ 
 % & & %\multicolumn{1}{l|}{\textbf{Control}} & 
  %\multicolumn{3}{l}{\textbf{Emotion}} & 
  %\multicolumn{3}{l}{\textbf{Variability}} & 
  % \multicolumn{3}{l}{\textbf{}}& 
  % \multicolumn{3}{l}{\textbf{Rate}}\\
 \hline
\textbf{ Dataset} &\textbf{MHC--Control} & % &\multicolumn{1}{l|}{Emotion Dimension} & 
 % \multicolumn{1}{l}{V} & \multicolumn{1}{l}{A} & \multicolumn{1}{l}{D} & 
 % \multicolumn{1}{l}{V} & \multicolumn{1}{l}{A} & \multicolumn{1}{l}{D} &
 \multicolumn{1}{c}{V} & \multicolumn{1}{c}{A} & \multicolumn{1}{c}{D} &
 \multicolumn{1}{c}{V} & \multicolumn{1}{c}{A} & \multicolumn{1}{c}{D} \\ \hline
 % & \multicolumn{1}{l|}{} & \multicolumn{1}{l|}{} &  & \multicolumn{1}{l|}{} & \multicolumn{1}{l|}{} &  & \multicolumn{1}{l|}{} & \multicolumn{1}{l|}{} &  & \multicolumn{1}{l|}{} & \multicolumn{1}{l|}{} &  \\
%& \multicolumn{1}{p{0.8cm}}{\textbf{V}} & & & & & \\
Twitter-STMHD & ADHD--control &  -- & -- & 0.001 & --  & 0.001 & 0.002\\ %& 0.018 & 0.021 & 0.030 & -0.008 & -0.007 & -0.006 % (-.011, -.006)
& Bipolar--control & -- & -- & -- & 0.002 & -- & --\\ %(-.018,-.011) %& -- & 0.013 & 0.029 & -0.015 & -0.007 & -0.008

& MDD--control & 0.004 & -- & -- & 0.004 & 0.003 & 0.005\\ %(-.023,-.012) %0.027 & -- & 0.059 & -0.017 & -0.009 & -0.009
& OCD--control & -- & -- & 0.002 &  -- & 0.002 & 0.002\\ %(-.0136,-.007) %-- & 0.023 & 0.033 & -0.010 & -0.007 & -0.005 &
& PPD--control &  -- & -- & -- &  -- & -- & --\\ %(-.013,.002) %-- & 0.046 & 0.051 &  -- & -0.006 & -0.006 
& PTSD--control & 0.002 & 0.001 & -- & 0.002 & 0.001 & 0.001 \\ %(-.018, -.013) %0.023 & -- & 0.018 & -0.015 & -0.007 & -0.006

& Depression--control & 0.002 & -- & 0.002 & 0.002 & 0.001 & 0.002 \\ %(-.015,-.011) %-- & 0.015 & 0.040 & -0.013 & -0.007 & -0.006
Reddit eRisk& \DT{Depression--control} & -- & -- & -- &  -- & -- & -- \\ % -- & -- & $\downarrow$ & $\uparrow$ & -- & $\uparrow$

    \hline
%\end{tabulary} 
\end{tabular}
}
% \vspace*{-1mm}
\caption{ % \textbf{Valence (V), Arousal (A), Dominance (D)}: 
The mean difference in UED metrics % across MHC groups compared to 
between each MHC group and the control if there was a significant difference between groups. 
E.g., $0.004$ for MDD--control and rise rate `V' %column 
means that the MDD group has significantly higher rise rate for valence by $0.004$ than the control group.
}
% \vspace*{-3mm}
\label{tab:mean_diff_rise_recover}
\end{table*}

\end{document}